\documentclass[journal,compsoc]{IEEEtran}
\usepackage{amsmath,amsfonts}
\usepackage{algorithmic}
\usepackage{algorithm}
\usepackage{array}
\usepackage[caption=false,font=normalsize,labelfont=sf,textfont=sf]{subfig}
\usepackage{textcomp}
\usepackage{stfloats}
\usepackage{color}
\usepackage{url}
\usepackage{verbatim}
\usepackage{graphicx}
\usepackage{cite}
\usepackage{multirow}
\usepackage{hyperref}
\hypersetup{
	colorlinks=true,
	linkcolor=red,
	urlcolor=blue,
}
\usepackage{multicol}
\usepackage{ragged2e}
\usepackage{caption}
\begin{document}

\title{A Comprehensive Review of Image Line Segment Detection and Description: Taxonomies, Comparisons, and Challenges}

\author{Xinyu Lin,~\IEEEmembership{Member,~IEEE} \\
	Yingjie Zhou,~\IEEEmembership{Member,~IEEE}, 
	Yipeng Liu,~\IEEEmembership{Senior Member,~IEEE}, 	
	and Ce Zhu*,~\IEEEmembership{Fellow,~IEEE}
	\newline
	\thanks{Xinyu Lin (roylin\_cv@163.com), Yipeng Liu, and Ce Zhu (* Corresponding author) are with the School of Information and Communication Engineering, University of Electronic Science and Technology of China (UESTC), Chengdu 611731, China. Yingjie Zhou is with the College of Computer Science, Sichuan University, Chengdu, Sichuan, China.}}

\IEEEtitleabstractindextext{
\justifying
\begin{abstract}
	An image line segment is a fundamental low-level visual feature that delineates straight, slender, and uninterrupted portions of objects and scenarios within images. Detection and description of line segments lay the basis for numerous vision tasks. Although many studies have aimed to detect and describe line segments, a comprehensive review is lacking, obstructing their progress. This study fills the gap by comprehensively reviewing related studies on detecting and describing two-dimensional image line segments to provide researchers with an overall picture and deep understanding. Based on their mechanisms, two taxonomies for line segment detection and description are presented to introduce, analyze, and summarize these studies, facilitating researchers to learn about them quickly and extensively. The key issues, core ideas, advantages and disadvantages of existing methods, and their potential applications for each category are analyzed and summarized, including previously unknown findings. The challenges in existing methods and corresponding insights for potentially solving them are also provided to inspire researchers. In addition, some state-of-the-art line segment detection and description algorithms are evaluated without bias, and the evaluation code will be publicly available. The theoretical analysis, coupled with the experimental results, can guide researchers in selecting the best method for their intended vision applications. Finally, this study provides insights for potentially interesting future research directions to attract more attention from researchers to this field.
\end{abstract}

}
\maketitle


\section{Introduction}
\label{sec_intro}
\begin{figure}[tbp]
	\centering
	\includegraphics[width=\linewidth]{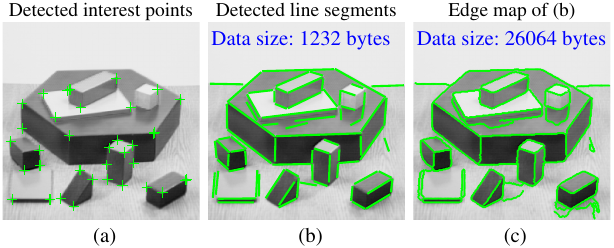}
	\caption{Local image features on a test image: (a) Interest points \cite{FASTER}, (b) Line segments \cite{EDLines}, and (c) Corresponding edge map \cite{EdgeDrawing} of line segments in (b). Line segments contain more structural and geometric features about a scene than points and are more compact in data expression than edges.}
	\label{point_line_edge_demo}
\end{figure}

\IEEEPARstart{A}{s} a typical two-dimensional (2D) low-level image feature \cite{local_feature}, line segments \cite{LSD} contain more structural and geometric features about the scene than points \cite{SIFT,FASTER,KeyNet}, as shown in Fig. \ref{point_line_edge_demo}. In addition, they have the advantage of compact data expression and strong orientation constraints, and can completely support multiview geometry theory \cite{MultipleViewGeometryinComputerVision} in both 2D and three-dimensional (3D) spaces, compared with edges \cite{Canny}. Because of these excellent characteristics, line segments are widely used in numerous vision tasks, such as visual measurement \cite{1512061}, camera pose estimation \cite{w-MSAC,our_demo,PoseEstimationfromLineCorrespondencesACompleteAnalysisandaSeriesofSolutions}, visual odometry (VO) \cite{8691513}, visual simultaneous mapping and localization (V-SLAM) \cite{9521742}, and structure from motion (SFM) \cite{8395064}. In these vision tasks, line segments lay the foundation for other processes by providing line segment matches in different images containing the same scene.

Generally, line segment matches are obtained through the following two steps. Line segments are first detected in images using developed detectors and then described by designed descriptors, which can be regarded as the signatures of line segments and matched by their descriptor similarity. Many line segment detection and description algorithms have been designed in recent decades. However, unlike well-reviewed points \cite{TPAMI_Interest_Point_Survey} and edges \cite{JING2022259}, no comprehensive review is available that systematically analyzes and summarizes them, making them difficult for researchers to fully visualize and deeply understand. Specifically, (1) it is difficult and  time-consuming for researchers to determine the complex relationship among these algorithms, their key issues, core ideas, advantages and disadvantages, and potential applications. Although many studies have been proposed to robustly detect and describe line segments from images, some common challenges and unsolved problems remain in this field. (2) Almost all these studies reported that their approaches perform the "best" regarding their designed experiments, making it difficult for researchers to select a suitable line segment detection and description method for their intended applications. (3) Furthermore, compared with well-studied point features, research on line segment detection and description is ongoing, and more interesting future research directions need to be explored.

\begin{figure*}[tbp]
	\centering
	\includegraphics[width=1\textwidth]{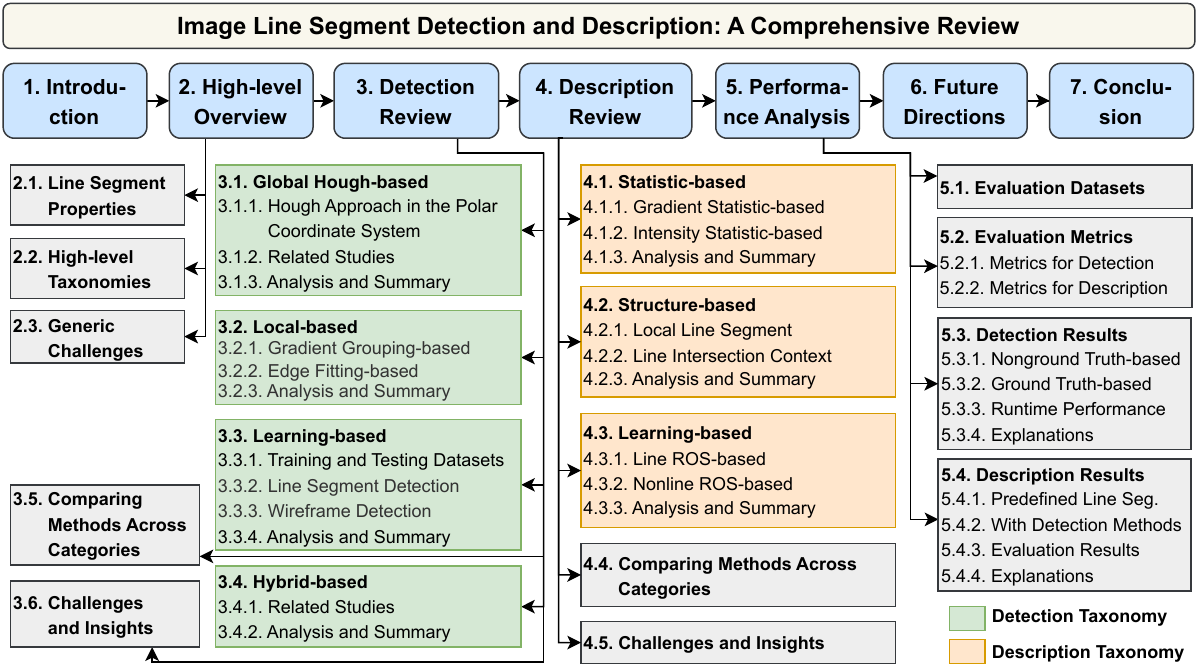}
	\caption{The structure of this review and high-level taxonomies of existing line segment detection and description methods.}
	\label{line_segment_taxonomy}
\end{figure*}

This study addresses the problems above by comprehensively reviewing numerous line segment detection and description algorithms to provide researchers with an overall picture and deep understanding of them. The contributions of this study are summarized as follows.

\begin{itemize}
	\item As shown in Fig. \ref{line_segment_taxonomy}, two taxonomies introducing and analyzing numerous line segment detection and description methods are presented according to their mechanisms, facilitating researchers to learn about these methods quickly and extensively.
	
	\item For the methods in each category, their key issues, core ideas, advantages and disadvantages, and potential applications are analyzed and summarized, including some previously unknown findings.
	
	\item The challenges in existing line segment detection and description methods and corresponding insights for potentially dealing with them are also provided to inspire researchers.
	
	\item Instead of task-specific evaluation \cite{EVOLIN}, a generic and unbiased performance evaluation is performed for various state-of-the-art (SOTA) line segment detection and description methods based on natural and synthetic datasets. The theoretical analysis, coupled with the quantitative experimental results, can guide researchers in selecting an appropriate line segment detection and description method for their intended applications. The evaluation code (\url{https://github.com/roylin1229/line_segment_review}) will be released publicly, facilitating researchers to evaluate their developed algorithms and test other methods.
	
	\item Since the research on line segment detection and description is in development, this study provides some insights for potentially interesting future research directions to attract more attention from researchers to this field.
\end{itemize}

The rest of this review is organized as shown in Fig. \ref{line_segment_taxonomy}. A high-level overview of line segment detection and description is introduced in Section \ref{sec_high_level_overview}. Section \ref{sec_line_detection} and Section \ref{sec_line_description} review numerous line segment detection and description methods, respectively. Numerical experiments on natural and synthetic datasets are performed in Section \ref{sec_performance}. Section \ref{research_directions} provides some insights for potentially interesting future research directions. Section \ref{sec_conclusion} summarizes the conclusions.

\section{High-level Overview of Line Segment Detection and Description}
\label{sec_high_level_overview}
An image line segment is a fundamental low-level visual element, defined as straight, slender and continuous sections of objects and scenarios within digital images. Line segments generally appear in image regions where the image feature has a trend and significant variance, as shown in Fig. \ref{point_line_edge_demo} (b). They are not dependent on specific objects and scenarios and are widely used in various vision tasks. They can be expressed by line segment endpoints or the combination of line segment midpoints, radii, and angles.

\subsection{Line Segment Properties}
\label{subsec_ls_properties}
As a typical low-level image feature \cite{local_feature}, line segments should have the following properties:
\begin{itemize}
	\item \textbf{Repeatability}: A fundamental attribute of line segments is their capacity for consistent, robust, and stable detection, description, and matching across multiple images depicting the same scene under varying imaging conditions, such as noise, occlusion, rotation, scale, illumination, and viewpoint changes. This attribute is indispensable for vision tasks necessitating the establishment of correspondences between line segments in such images.
	
	\item \textbf{Distinctiveness}: The second line segment property is the ability to be uniquely identified and distinguished from other features in an image.
	
	\item \textbf{Accuracy}: The line segments should be detected accurately in an image, including the location of endpoints, related orientations, and scales.
	
	\item \textbf{Efficiency}: The detection and description of line segments should be efficient regarding computational complexity and memory usage since they serve as the foundation for numerous vision tasks, particularly in resource-limited systems and platforms.
	
	\item \textbf{Quantity}: The detected and matched line segments (based on line segment description) should satisfy the requirements of specific vision tasks.
\end{itemize}

Line segments typically contain more structural and geometric information about a scene than point features. Consequently, their locality \cite{local_feature} may be slightly compromised compared to point features, as they are often detected within a larger image region.

\subsection{High-level Taxonomies}
\label{subsec_high_level_taxonomy}
As mentioned above, line segments form the basis for vision tasks by establishing matches across different images, typically achieved in two critical steps: first detecting and then describing line segments. After decades of development, numerous line segment detection and description algorithms have been proposed. According to the mechanism of existing line segment detection methods, they can be roughly classified into four categories, as shown in Fig. \ref{line_segment_taxonomy}.

\begin{enumerate}
	\item \textbf{Global Hough-based} methods \cite{METHODANDMEANSFORRECOGNIZINGCOMPLEXPATTERNS,UseoftheHoughTransformationtoDetectLinesandCurvesinPictures,GeneralizingtheHoughtransformtodetectarbitraryshapes,35497,HoughTransformfromtheRadonTransfor,CompletelinesegmentdescriptionusingtheHoughtransform,AfastHoughtransformforsegmentdetection,Houghtransformmodifiedbylineconnectivityandlinethickness,CompletedescriptionofmultiplelinesegmentsusingtheHoughtransform,RobustDetectionofLinesUsingtheProgressiveProbabilisticHoughTransform,AccurateandRobustLineSegmentExtractionbyAnalyzingDistributionaroundPeaksinHoughSpace,ExtendedHoughtransformforlinearfeaturedetection,OnStraightLineSegmentDetection,AnImprovedHoughTransformNeighborhoodMapforStraightLineSegments,CollinearSegmentDetectionUsingHTNeighborhoods,Connectivity-EnforcingHoughTransformfortheRobustExtractionofLineSegments,AnAccurateMethodforLineDetectionandManhattanFrameEstimation,AStatisticalMethodforLineSegmentDetection,AccurateandRobustLineSegmentExtractionUsingMinimumEntropyWithHoughTransform,ClosedFormLineSegmentExtractionUsingtheHoughTransform,Bachiller2017,PClines} exploit the duality between points and curves in the image and Hough spaces to detect line segments, in which the line segment detection problem in the image space is transformed into a peak statistic problem in the Hough space. (subsection \ref{subsec_detection_hough})
	
	\item \textbf{Local-based} methods with subcategories of gradient grouping-based and edge fitting-based approaches leverage the consistent image features in local regions to detect line segments, followed by merging and validation processes to improve line segment quality. \textbf{Gradient grouping-based} methods \cite{ExtractingStraightLines,Token-basedextractionofstraightlines,Findinglinesegmentsbystickgrowing,LSD,LSDaLineSegmentDetector,Linesegmentdetectionusingweightedmeanshiftproceduresona2Dslicesamplingstrategy,AParameterlessLineSegmentandEllipticalArcDetectorwithEnhancedEllipseFitting,MultiscalelinesegmentdetectorforrobustandaccurateSfM,JointAContrarioEllipseandLineDetection,ANovelLineletBasedRepresentationforLineSegmentDetection,FSG,PLSD,AG3line,FastLineSegmentDetectionandLargeSceneAirportDetectionforPolSAR,ALineSegmentDetectorforSpaceTargetImagesRobusttoComplexIllumination} group image regions with highly consistent gradient orientations or magnitudes to detect line segments. \textbf{Edge fitting-based} methods \cite{Extractionofstraightlinesinaerialimages,ASimpleandRobustLineDetectionAlgorithmBasedonSmallEigenvalueAnalysis,EDLines,Robustlinedetectionusingtwoorthogonaldirectionimagescanning,Outdoorplacerecognitioninurbanenvironmentsusingstraightlines,CannyLines,OTLines,LB-LSD,PropertySimilarityLineSegmentDetector,ELSED,Ultrafastlinedetector, E2LSD,E2LSD-J,ANovelPixelOrientationEstimationBasedLineSegmentDetectionFrameworkandItsApplicationstoSARImages} detect line segments by fitting highly consistent edge points. (subsection \ref{subsec_local_based})

	\item \textbf{Learning-based} methods with \textbf{line segment} \cite{Sem-LSD,LearningAttractionFieldRepresentationforRobustLineSegmentDetection,LearningtoCalibrateStraightLinesforFisheyeImageRectification,TP-LSD,LGNN,L2D2,FullyConvolutionalLineParsing,TowardsRealtimeandLightweightLineSegmentDetection,ULSD,SOLD2,Superline,LineSegmentDetectionUsingTransformerswithoutEdges,LearningRegionalAttractionforLineSegmentDetection,ELSD,SelfsupervisedLightweightLineSegmentDetectorandDescriptor} and \textbf{wireframe} (with more than one line segment and the corresponding junction) \cite{LearningtoParseWireframesinImagesofMan-MadeEnvironments,WireframeParsingWithGuidanceofDistanceMap,PPGNet,End-to-EndWireframeParsing, LearningtoReconstruct3DManhattanWireframesFromaSingleImage,Holistically-AttractedWireframeParsing,HAWP,Hole-robustWireframeDetection,SemanticRoomWireframeDetectionfromaSingleView} detection subcategories learn the line segment or wireframe patterns by designing networks and loss functions in carefully prepared training and testing datasets. (subsection \ref{subsec_detection_learning})
		
	\item \textbf{Hybrid-based} methods \cite{MCMLSD,DeepHoughTransformLinePriors,LinedetectionviaalightweightCNNwithaHoughlayer,LSDNet,DeepLSD} incorporate multiple mechanisms to detect image line segments. The fundamental idea behind these approaches is to amalgamate the strengths of multiple mechanisms to overcome the drawbacks of the specific mechanism. These approaches provide a high degree of flexibility for customization. (subsection \ref{subsec_hybrid})
\end{enumerate}

Existing line segment description methods can be roughly classified into three categories according to the mechanism, as shown in Fig. \ref{line_segment_taxonomy}.
\begin{enumerate}
	\item \textbf{Statistic-based} methods with \textbf{gradient statistic-based} \cite{HLD,MSLD,Extendpointdescriptorsforlinecurveandregionmatching,Anoveldescriptorforline(curve)matching,FastLineDescriptionforLinebasedSLAM,Matchingofstraightlinesegmentsfromaerialstereoimagesofurbanareas,LineMatchingUsingAppearanceSimilaritiesandGeometricConstraints,LBD,Scaleinvariantlinedescriptorsforwidebaselinematching,Arobustlinematchingmethodbasedonlocalappearancedescriptorandneighboringgeometricattributes,RemoteSensingImageRegistrationwithLineSegmentsandTheirIntersections,LineSegmentMatchingFusingLocalGradientOrderandNonLocalStructureInformation,ALineDescriptorforSLAM,IILB} and \textbf{intensity statistic-based} subcategories \cite{Widebaselinestereomatchingwithlinesegments,Scaleinvariantlinedescriptorsforwidebaselinematching,Twoviewlinematchingalgorithmbasedoncontextandappearanceinlowtexturedimages,Arobustlinematchingmethodbasedonlocalappearancedescriptorandneighboringgeometricattributes,ALineMatchingMethodBasedonMultipleIntensityOrderingwithUniformlySpacedSampling,LineSegmentMatchingFusingLocalGradientOrderandNonLocalStructureInformation,IILB} exploit the statistic features, \textit{e.g.}, the histogram of gradients, or pixel intensities in line regions of support (ROSs) to encode line segments. A few of them \cite{Scaleinvariantlinedescriptorsforwidebaselinematching,Arobustlinematchingmethodbasedonlocalappearancedescriptorandneighboringgeometricattributes,LineSegmentMatchingFusingLocalGradientOrderandNonLocalStructureInformation,IILB} leverage both gradients and intensities to formulate descriptors. (subsection \ref{subsec_description_statistic})
		
	\item \textbf{Structure-based} methods with \textbf{local line segment-assisted} \cite{Linepatternretrievalusingrelationalhistograms,WidebaselineimagematchingusingLineSignatures,Robustaffineinvariantlinematchingforhighresolutionremotesensingimages,Anovelsimilarityinvariantlinedescriptorforgeometricmapregistration,NovelSimilarityInvariantLineDescriptorandMatchingAlgorithmforGlobalMotionEstimation,Twoviewlinematchingalgorithmbasedoncontextandappearanceinlowtexturedimages,AutomaticRegistrationMethodforOpticalRemoteSensingImageswithLargeBackgroundVariationsUsingLineSegments,MultimodalImageRegistrationWithLineSegmentsbySelectiveSearch} and \textbf{line intersection context-assisted} subcategories \cite{Anovellinematchingmethodbasedonintersectioncontext,Simultaneouslinematchingandepipolargeometryestimationbasedontheintersectioncontextofcoplanarlinepairs,LineMatchinginWideBaselineStereoATopDownApproach,Widebaselinestereomatchingbasedonthelineintersectioncontextforrealtimeworkspacemodeling,RobustLineMatchingBasedonRay-Point-RayStructureDescriptor,HierarchicalLineMatchingBasedonLineJunctionLineStructureDescriptorandLocalHomographyEstimation,Linesegmentmatchingandreconstructionviaexploitingcoplanarcues} use additional local line segments to structurally describe the relative relationships among them. (subsection \ref{subsec_description_structure})
		
	\item \textbf{Learning-based} methods with \textbf{line ROS-based} \cite{Abinaryrobustlinedescriptor,DLD,WLD,L2D2,Towardslearninglinedescriptorsfrompatchesanewparadigmandlargescaledataset} and \textbf{nonline ROS-based} subcategories \cite{LearnableLineSegmentDescriptorforVisualSLAM,SOLD2,Superline,LineasaVisualSentenceContextAwareLineDescriptorforVisualLocalization,ELSD,LDAM} learn the line segment representations locally or globally based on designed networks with loss functions regarding prepared training and testing datasets. (subsection \ref{subsec_description_learning})
\end{enumerate}

Note that multiple line segment descriptors can be combined to improve the description ability, as in \cite{Twoviewlinematchingalgorithmbasedoncontextandappearanceinlowtexturedimages}. Additionally, according to the data expression, these descriptors can be classified into binary and float descriptors. Except for the descriptors in \cite{Abinaryrobustlinedescriptor,IILB} and the binary version of \cite{LBD} in OpenCV, almost all other line segment descriptors are float descriptors.

\subsection{Generic Challenges}
\label{subsec_generic_challenges}
Despite numerous studies aimed at detecting and describing line segments from images, certain generic challenges remain in this field. Specifically,
\begin{itemize}
	\item \textbf{Unstable endpoints of line segments}. The first general challenge for line segment detection and description is that the endpoints of detected line segments are generally unstable due to many factors, such as image noise, occlusion, and illumination variance. This makes the line segment description and matching difficult since the corresponding line ROS or neighborhood relationship among line segments is generated according to the line segment endpoints.
		
	\item \textbf{Imperfect detection and description methods}. The objective of line segment detection and description algorithms is to exhibit the diverse properties mentioned in subsection \ref{subsec_ls_properties} that line segments are expected to possess. However, it is frequently observed that numerous algorithms demonstrate exceptional performance in only one or some aspects. Therefore, discovering line segment detection and description algorithms that can simultaneously meet the multiple property requirements of line segments, such as accuracy, repeatability, robustness, and efficiency, mentioned in subsection \ref{subsec_ls_properties}, is challenging.
		
	\item \textbf{Unsatisfactory performance in challenging scenarios.} Line segment detection and description usually perform unsatisfactorily in extremely challenging scenarios, including those with significant changes in illumination, wide variations in viewpoint, low-texture scenes, and highly repetitive environments.
	
	\item \textbf{Shortage of gold-standard datasets.} The availability of gold-standard datasets is critical for advancing the research in line segment detection and description. However, current datasets with ground truth manually labeled with human expertise \cite{LearningtoParseWireframesinImagesofMan-MadeEnvironments,ANovelLineletBasedRepresentationforLineSegmentDetection,YorkUrban} suffer from limitations such as scale, consistency, completeness, correctness, and accuracy in annotations, as discussed in \cite{LSDNet}. Moreover, existing evaluation datasets with geometric constraints \cite{vggaffine,HPatches,KADID,DNIM,RDNIM} primarily concentrate on limited image variances, making it challenging to simulate real-world imaging processing scenarios effectively.
\end{itemize}

\section{Line Segment Detection Review}
\label{sec_line_detection}
This section reviews numerous line segment detection methods. As introduced above, these detection methods can be roughly classified into four categories: (1) global Hough-based methods (subsection \ref{subsec_detection_hough}), (2) local-based methods (subsection \ref{subsec_local_based}) with subcategories of gradient grouping-based and edge fitting-based approaches, (3) learning-based methods (subsection \ref{subsec_detection_learning}) with line segment and wireframe detection subcategories, and (4) hybrid methods (subsection \ref{subsec_hybrid}). First, this section introduces a subset of notable approaches from each category. Subsequently, an analysis and summary of the algorithms within each category are presented to provide a thorough understanding of them.

These detection methods have advantages, disadvantages, and positive applications, as summarized in TABLE \ref{line_segment_detection_comparsion} and analyzed in subsection \ref{subsec_detection_pro_cons}. The insights for potentially addressing the challenges in these methods are presented to inspire researchers, as detailed in subsection \ref{subsec_detection_challenges}. TABLE \ref{line_detection_implementation} lists some open-source implementations of these methods. 

\subsection{Global Hough-based Approaches}
\label{subsec_detection_hough}

\begin{figure}[tbp]
	\centering
	\includegraphics[width=0.49\textwidth]{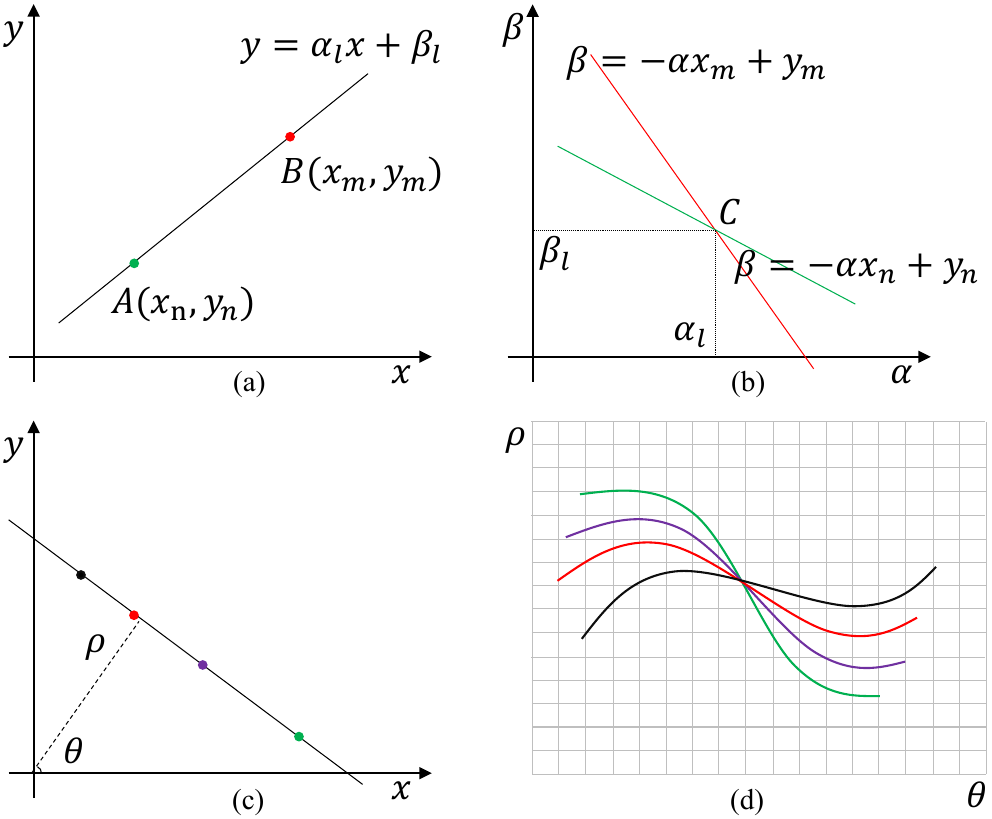}
	\caption{(a) - (b): The example in \cite{ExtendedHoughtransformforlinearfeaturedetection} shows the ideal Hough transformation in the Cartesian coordinate system. Two points $A(x_n, y_n)$ and $B(x_m, y_m)$ on a line $y=\alpha_lx+\beta_l$ in the image space correspond to two lines $\beta=-\alpha x_n + y_n$ and $\beta=-\alpha x_m + y_m$ in the Hough space, and they intersect at a point $C(\alpha_l, \beta_l)$. (c) - (d): The example shows the ideal Hough transformation in the polar coordinate system. Four points on a line in the image space correspond to four curves in the Hough space, and they intersect at a point.}
	\label{hough_line}
\end{figure}

\subsubsection{Hough Approach in the Polar Coordinate System}
Hough first proposed the Hough transformation \cite{METHODANDMEANSFORRECOGNIZINGCOMPLEXPATTERNS} and applied it to line detection. In the Cartesian coordinate system, a line $y=\alpha x + \beta$ in the image space corresponds to a point $(\alpha, \beta)$ in the Hough space, and vice versa. As shown in Fig. \ref{hough_line} (a) and (b), some points on a line in the image space correspond to some lines in the Hough space, and they intersect at a point. The lines can be detected by searching for the intersection point in the Hough space.

The approach in \cite{METHODANDMEANSFORRECOGNIZINGCOMPLEXPATTERNS} has a singularity problem. When lines in the image space are orthogonal to the $x$-axis, there are no intersecting points in the Hough space, meaning that the vertical lines cannot be detected. To address this issue, Duda et al. utilized the polar coordinate system to replace the Cartesian coordinate system for the Hough transformation \cite{UseoftheHoughTransformationtoDetectLinesandCurvesinPictures}, eliminating the singularity problem. A point in the image space can be transformed into a curve in the Hough space by leveraging the polar coordinate transformation $\rho = x cos \theta + y sin \theta$ and vice versa. As shown in Fig. \ref{hough_line} (c) and (d), some points on a line in the image space correspond to some curves in the Hough space, intersecting with each other at a point. Similarly, the lines can be detected by searching the intersecting point (peak) in the Hough space.

\subsubsection{Related Studies}
Based on \cite{UseoftheHoughTransformationtoDetectLinesandCurvesinPictures}, many related studies have been proposed to improve the line (segment) detection performance, as surveyed in \cite{AsurveyofHoughTransform,Rahmdel2015ARO}. The study in \cite{CompletelinesegmentdescriptionusingtheHoughtransform} and its improved version in \cite{CompletedescriptionofmultiplelinesegmentsusingtheHoughtransform} were presented to efficiently and effectively locate line endpoints. Matas presented a progressive probabilistic Hough transformation to improve the computational efficiency and decrease the memory cost \cite{RobustDetectionofLinesUsingtheProgressiveProbabilisticHoughTransform} for line segment detection. By analyzing the distribution around peaks in the Hough space, Furukawa et al. presented a robust line segment detection method \cite{AccurateandRobustLineSegmentExtractionbyAnalyzingDistributionaroundPeaksinHoughSpace}. Du et al. presented a collinear line segment detection approach by exploiting the information of wings around the peaks in the Hough space \cite{CollinearSegmentDetectionUsingHTNeighborhoods}. Based on parallel coordinates, the authors in \cite{PClines} proposed an alternative parameterization for the Hough transform to detect line segments. The authors in \cite{AnAccurateMethodforLineDetectionandManhattanFrameEstimation} presented the concept of "soft voting" in the Hough space, which leverages the uncertainty information in edge detection to improve detection performance. Xu et al. detected line segments based on minimum entropy analysis \cite{AccurateandRobustLineSegmentExtractionUsingMinimumEntropyWithHoughTransform}. In addition, they presented a statistical method for line segment detection by considering various information, such as quantization errors \cite{AStatisticalMethodforLineSegmentDetection}.

\subsubsection{Analysis and Summary}
\textit{\textbf{The core idea of global Hough-based approaches lies in exploiting the duality between points and curves in the image and Hough spaces to detect line segments, in which the line segment detection problem in the image space is transformed into a peak statistic problem in the Hough space.}} Therefore, analyzing and counting the peaks in the Hough space is critical. Considering the computational and memory costs, the quantitative values need to be utilized in searching for the intersection points (peaks) in the Hough space, \textit{e.g.}, the quantitative values of $\rho$ and $\theta$ in Fig. \ref{hough_line} (d). These quantitative values control the number and quality of detected lines or line segments to some extent. Furthermore, since images generally contain many pixels, performing the Hough transformation for each pixel is unacceptable, particularly for applications with limited computational and memory resources. Generally, input images need to be processed in advance to eliminate unnecessary processing pixels. A general approach is performing binary processing, such as edge detection \cite{Canny}. In this way, only these selected pixels are transformed into Hough space to further detect line segments, saving computational and memory resources.

\subsection{Local-based Approaches}
\label{subsec_local_based}
Local-based approaches leverage the consistent image features in local regions to detect line segments. According to the selection of image features, they can be further classified into gradient grouping-based and edge fitting-based approaches using image gradients and edges, respectively.

\subsubsection{Gradient Grouping-based Approaches}
\label{subsubsec_detection_gradient}

\begin{figure}[tbp]
	\centering
	\includegraphics[width=0.49\textwidth]{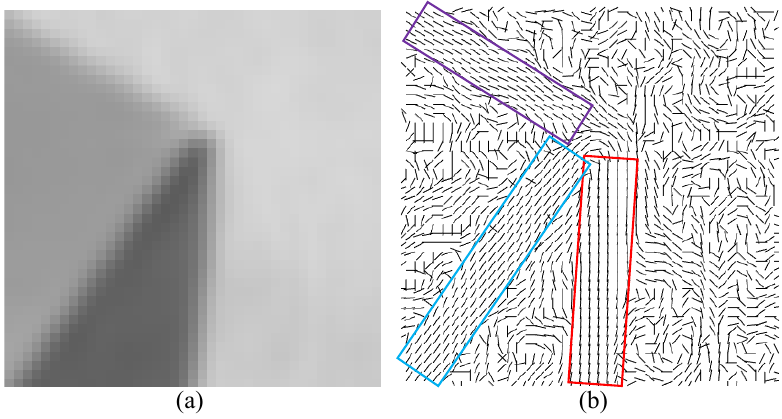}
	\caption{Visual illustrations in the LSD \cite{LSD,LSDaLineSegmentDetector} algorithm: (a) image region of interest; (b) level lines (small black lines) and the approximated rectangles for line ROSs. Line segments are detected from these approximated rectangles.}
	\label{lsd_level_line}
\end{figure}

\textbf{LSD Method.} The baseline method for line segment detection is the LSD method \cite{LSD, LSDaLineSegmentDetector}. Its mechanism is straightforward but effective. As shown in Fig. \ref{lsd_level_line} (b), the level lines orthogonal to gradient orientations for image pixels are calculated to group a series of line ROSs. Based on these line ROSs, the corresponding approximated rectangles around them are obtained and validated based on the aligned points and total points inside them using a \textit{contrario} and Helmholtz model \cite{desolneux2007gestalt}. The line segment parameters are calculated based on the approximated rectangle if valid.

\textbf{Related Studies.} In addition to the LSD method, other algorithms have grouped the image gradient for line segment detection. The authors in \cite{AParameterlessLineSegmentandEllipticalArcDetectorwithEnhancedEllipseFitting} and \cite{JointAContrarioEllipseandLineDetection} utilized image gradient orientations to group a series of line ROSs and extracted line segments from these line ROSs. By default, the LSD method was developed on a signal scale. The author in \cite{MultiscalelinesegmentdetectorforrobustandaccurateSfM} extended the LSD method by detecting line segments at multiple scales. Cho et al. first exploited the gradients in the horizontal and vertical directions to find these local maximum candidates for line segments and then detected, grouped, estimated, validated, and aggregated line segments based on these local maximum candidates \cite{ANovelLineletBasedRepresentationforLineSegmentDetection}. Considering that the line segments detected by the LSD method may contain small line fragments, the authors in \cite{FSG, PLSD} further grouped them into long line segments.

\subsubsection{Edge Fitting-based Approaches}
\label{subsubsec_detection_edge}
\begin{figure}[tbp]
	\centering
	\includegraphics[width=0.49\textwidth]{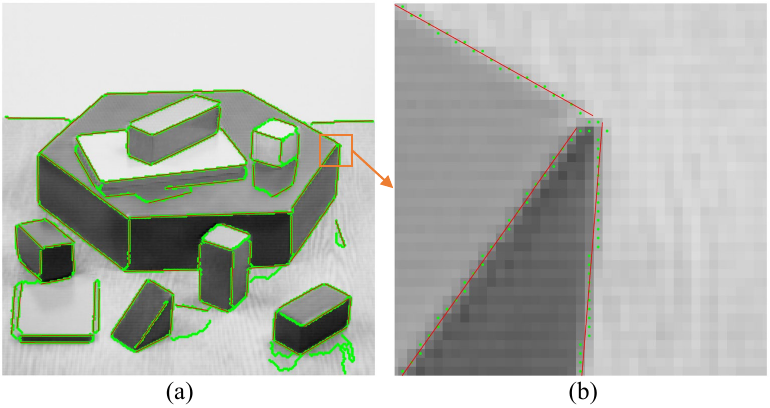}
	\caption{In the EDLines \cite{EDLines} algorithm, line segments (red lines) are detected from edges (green points). The figure shows a test image (a) and a region of interest (b) with drawn edges and detected line segments.}
	\label{edlines}
\end{figure}

\textbf{EDLines Method.} In addition to the LSD method, another baseline method is the EDLines \cite{EDLines} method, which exploits edges for line segment detection. The EDLines algorithm is reported to be 10+ times faster than the LSD algorithm because line segments are extracted only from these extracted edges rather than from whole images. In the EDLines method, edge chains are extracted by an ultrafast edge drawing \cite{EdgeDrawing} algorithm. As shown in Fig. \ref{edlines}, line segments are detected by fitting the edge points using least squares and validated using the Helmholtz criteria \cite{desolneux2007gestalt}.

\textbf{Related Studies.} In addition to EDLines, other methods have employed image edges for line segment detection. The authors in \cite{Outdoorplacerecognitioninurbanenvironmentsusingstraightlines} extracted line segments from Canny edges by continually fitting line segments until the predefined conditions were satisfied. Lu et al. proposed a parameter-free Canny operator to extract image edges and then detected line segments from these edges \cite{CannyLines}. The authors in \cite{ELSED} presented an enhanced line segment drawing (ELSED) method for line segment detection. It was reported as the fastest line segment detection algorithm to date. The authors in \cite{E2LSD,E2LSD-J} argued that line segments should be located on edge points with consistent coordinates and level lines.

\subsubsection{Analysis and Summary}
Local-based methods leverage consistent image features, \textit{e.g.}, gradients and edges, in local regions to detect line segments, followed by merging and validation processes to improve line segment quality. \textbf{\textit{The core idea of gradient grouping-based approaches lies in grouping image regions with highly consistent gradient orientations or magnitudes and further detecting corresponding line segments.}} Therefore, efficiently and effectively grouping these highly consistent region candidates in images is critical. Since most gradient grouping-based approaches only consider local connectivity for line segment detection and lack adequate semantic and global feature constraints, the region candidates are easily interrupted by noise and small occlusion. The detected line segments tend to be fragmentary instead of long and complete line segments; thus, additional processes, such as validation \cite{LSDaLineSegmentDetector} and aggregation \cite{ANovelLineletBasedRepresentationforLineSegmentDetection,Ametricforlinesegments}, are needed to robustly extract line segments. All of these processes control the quality of line segments to an extent. The detected image line segments generally have similar gradient properties in gradient magnitudes or orientations.

\textbf{\textit{The core idea of edge fitting-based approaches lies in fitting edge points based on least squares or other methods \cite{Whatsinasetofpointsstraightlinefitting,232082}, followed by merging \cite{Ametricforlinesegments} and validation \cite{EDLines} processes to improve line segment quality.}} Therefore, reliable and efficient edge extraction is critical for these approaches, particularly those based on edge chains, as in \cite{EDLines,ELSED,E2LSD,E2LSD-J}. The extracted edge quality and the aforementioned processes, such as aggregation and merging, deeply affect line segment detection performance.

\subsection{Learning-based Approaches}
\label{subsec_detection_learning}
\subsubsection{Widely Used Training and Testing Datasets}
\label{subsubsec_detection_dataset}
\begin{figure}[tbp]
	\centering
	\includegraphics[width=0.49\textwidth]{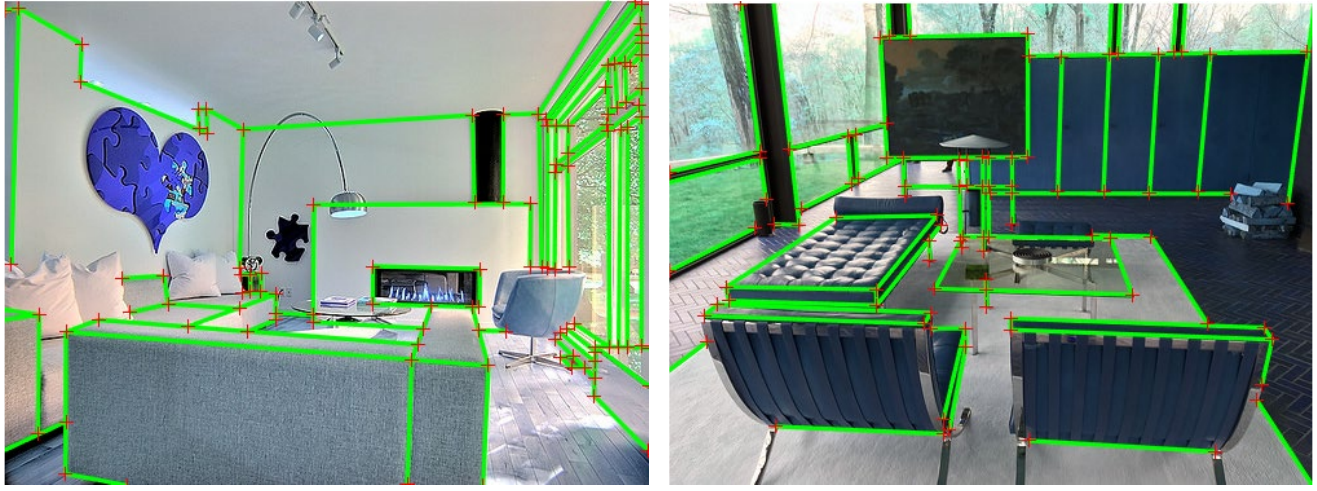}
	\caption{Labeled ground truth of line segments and junctions in the Wireframe \cite{LearningtoParseWireframesinImagesofMan-MadeEnvironments} dataset.}
	\label{Wireframes}
\end{figure}

Training and testing data lay the foundation for learning-based line segment detection approaches. The most widely used training and testing dataset is the Wireframe dataset \cite{LearningtoParseWireframesinImagesofMan-MadeEnvironments}, which includes 5,000 training images and 462 testing images with manually labeled ground truth of line segments and junctions. Wireframes are combined features comprising more than one line segment and corresponding junctions, as shown in Fig. \ref{Wireframes}. Therefore, the Wireframe dataset can be used in line segment detection and wireframe parsing tasks. Another popular dataset is YorkUrban \cite{YorkUrban, ANovelLineletBasedRepresentationforLineSegmentDetection}, which contains 102 testing images with manually labeled ground truth of Manhattan line segments. Due to its limited data size, many studies only used YorkUrban for evaluation.

\subsubsection{Related Studies on Line Segment Detection}
The authors in \cite{TP-LSD} trained a fast line segment detector, achieving real-time performance with the help of powerful GPU devices, in which the branch for line segment detection segments the pixels on straight lines from the background. Dai et al. developed a one-stage, fully convolutional line parsing network for line segment detection by predicting the center, position, length, and angle of line segments \cite{FullyConvolutionalLineParsing}. The authors in \cite{TowardsRealtimeandLightweightLineSegmentDetection} developed a lightweight line segment detection algorithm by minimizing a backbone network. In addition to performing only line segment detection, the authors in \cite{L2D2}, \cite{SOLD2}, \cite{Superline}, \cite{ELSD}, and \cite{SelfsupervisedLightweightLineSegmentDetectorandDescriptor} trained networks to simultaneously detect and describe line segments.

\subsubsection{Related Studies on Wireframe Detection}
Huang et al. presented the Wireframe dataset and trained two neural networks for wireframe detection \cite{LearningtoParseWireframesinImagesofMan-MadeEnvironments}, in which the branch network for line segment detection was designed by predicting whether a pixel falls on a line. In addition, they presented another wireframe parsing algorithm with the guidance of a distance map \cite{WireframeParsingWithGuidanceofDistanceMap}. Zhou et al. trained a network to detect wireframes directly from images \cite{End-to-EndWireframeParsing}. Xue et al. also presented an end-to-end network consisting of generation, matching, and verification components for line segments and their corresponding junctions \cite{Holistically-AttractedWireframeParsing,HAWP}.

\subsubsection{Analysis and Summary}
\label{subsec_detection_learning_analysis}
\textbf{\textit{The core idea of learning-based approaches lies in learning the line segment or wireframe patterns by designing networks and loss functions in carefully prepared training and testing datasets, in which both local and global image features in multiple layers can be fused to detect line segments.}} Therefore, designing networks regarding the prepared training and testing datasets is critical. The network generalization ability needs to be further validated since most networks are trained using the Wireframe dataset with only 5,000 samples and validated based on the limited testing datasets mentioned above. Some works tried to improve the generalization ability through self-supervision \cite{SOLD2}, ground truth line segments extracted from a point cloud \cite{L2D2}, and bootstrapping existing line detectors \cite{DeepLSD}.  Learning-based approaches tend to detect semantically meaningful line segments, such as those in indoor scenarios. In addition, similar to point features \cite{S2DNet}, they struggle to detect highly accurate line segments. One reason lies in that images are often downsized during training. In practice, they can also be trained with data augmentation and on challenging images to make them more robust to hard conditions and image noise. With the help of GPU devices, a few of these approaches can achieve real-time detection performance.

\subsection{Hybrid-based Approaches}
\label{subsec_hybrid}
\subsubsection{Related Studies}
Several hybrid-based approaches incorporating multiple mechanisms were recently presented to detect image line segments. Almazan et al. presented a Markov chain marginal line segment detector (MCMLSD) \cite{MCMLSD} by leveraging global probabilistic Hough transformation and perceptual grouping in the image domain. The authors in \cite{DeepHoughTransformLinePriors, LinedetectionviaalightweightCNNwithaHoughlayer} introduced the Hough transformation into neural networks, enabling the constructed network to extract global and local features for line segment detection. By leveraging a lightweight convolutional neural network with a classical LSD detector, a hybrid-based line segment detection was presented in \cite{LSDNet}, named LSDNet. Pautrat et al. proposed the DeepLSD method \cite{DeepLSD}, in which a line attraction field based on a designed neural network was generated and fed to a traditionally handcrafted line detector.

\subsubsection{Analysis and Summary}
\textbf{\textit{Hybrid-based approaches utilize a combination of multiple mechanisms to detect line segments in images. The fundamental idea behind these approaches is to amalgamate the strengths of multiple mechanisms to overcome the drawbacks of the specific mechanism.}} Therefore, the primary goal is to seamlessly integrate these mechanisms and design dedicated algorithms for line segment detection in relevant applications. In doing so, customizing the adopted mechanisms as per the situation is crucial, with all such modifications influencing the line segment detection quality.

\subsection{Comparing Methods Across Categories}
\label{subsec_detection_pro_cons}
Line segment detection methods in different categories have advantages, disadvantages, and positive applications, as summarized in TABLE \ref{line_segment_detection_comparsion}. TABLE \ref{line_detection_implementation} lists some open-source implementations of these methods. Global Hough-based approaches can effectively incorporate global image features to detect line segments. Thus, they are robust to image noise, occlusion, and small line segment fragments. In addition, they are mathematically explainable. However, most global Hough-based approaches lack adequate local connectivity and semantic feature constraints and thus may lead to an inaccurate result. They are suitable for vision applications requiring the detection of long line segments, such as horizon line \cite{10.1145/2964284.2967198} and vanishing point \cite{10.1007/978-3-030-01249-6_20} detection.

\begin{table*}[tbp] 	
	\centering 	
	\scriptsize 	
	\caption{Qualitative comparisons of line segment detection algorithms from the four categories.}		 	\begin{tabular}{|p{0.06\textwidth}<{\centering}|p{0.1\textwidth}<{\centering}|p{0.26\textwidth}<{\centering}|p{0.28\textwidth}<{\centering}|p{0.18\textwidth}<{\centering}|}
		\hline 		\textbf{Category}       & \textit{\textbf{e.g.}} & \textbf{Advantages}                                                                                                                                  & \textbf{Disadvantages}                                                                                                                      & \textbf{Typical Applications}                                \\ \hline 		Global Hough-based    & Hough\cite{UseoftheHoughTransformationtoDetectLinesandCurvesinPictures}, HoughP \cite{RobustDetectionofLinesUsingtheProgressiveProbabilisticHoughTransform}         & (1) Robust to image noise and occlusion by fusing global features, (2) Fewer line fragments, (3) Mathematically explainable. & (1) With abnormal line segments, (2) Fewer local connectivity and semantic feature constraints.                    & Vanishing point detection \cite{10.1007/978-3-030-01249-6_20}, Horizon line detection \cite{10.1145/2964284.2967198} \\ \hline 		Local-based & LSD \cite{LSD}, Linelet \cite{ANovelLineletBasedRepresentationforLineSegmentDetection}, EDLines \cite{EDLines}                   & (1) Low computational cost, (2) Good local connectivity constraint, (3) Mathematically explainable.                                                  & (1) Sensitive to image noise and occlusion, (2) With line fragments, (3) Fewer global and semantic feature constraints.          & Visual odometry \cite{8691513}, Visual localization \cite{PoseEstimationfromLineCorrespondencesACompleteAnalysisandaSeriesofSolutions}, Visual SLAM \cite{9521742} \\ \hline 		Learning-based & MLSD \cite{TowardsRealtimeandLightweightLineSegmentDetection}, LCNN \cite{End-to-EndWireframeParsing}, FClip \cite{FullyConvolutionalLineParsing}              & (1) End-to-end manner, (2)   Fusing local and global features for semantic line segment detection.                                      & (1) Deeply affected by training and testing data, (2) Difficult to explain mathematically, (3) High computational cost, needing GPU. & 3D reconstruction \cite{LearningtoReconstruct3DManhattanWireframesFromaSingleImage}, Image partitioning \cite{7298931}, Scene parsing \cite{6619245}                   \\ \hline 		Hybrid-based    & MCMLSD \cite{MCMLSD}, HTLCNN \cite{DeepHoughTransformLinePriors}         & Fusing multiple mechanisms to detect line segments. &  Depending on the specific fusion mechanism.                   & Depending on the specific fusion mechanism. \\ \hline		 	
	\end{tabular} 	
	\label{line_segment_detection_comparsion} 
\end{table*}

\begin{table}[tbp]
	\centering
	\scriptsize
	\setlength\tabcolsep{2pt}
	\caption{Open-source implementations of sampled line segment detection methods, in which the methods colored with blue are compared in this review.} 
	\begin{tabular}{|p{0.1\textwidth}<{\centering}|p{0.37\textwidth}<{\centering}|}
		\hline
		\textbf{Method} & \textbf{URL} \\
		\hline
		Hough & \url{https://opencv.org/} \\
		\hline
		\textcolor{blue}{HoughP} \cite{RobustDetectionofLinesUsingtheProgressiveProbabilisticHoughTransform} & \url{https://opencv.org/} \\
		\hline
		AG3line \cite{AG3line} & \url{https://github.com/weidong-whu/AG3line} \\
		\hline
		FSG \cite{FSG} & \url{https://github.com/iago-suarez/FSG} \\
		\hline
		\textcolor{blue}{Linelet} \cite{ANovelLineletBasedRepresentationforLineSegmentDetection} & \url{https://github.com/NamgyuCho/Linelet-code-and-YorkUrban-LineSegment-DB} \\
		\hline
		ELSDc \cite{JointAContrarioEllipseandLineDetection} & \url{https://github.com/viorik/ELSDc} \\
		\hline
		MLSD \cite{MultiscalelinesegmentdetectorforrobustandaccurateSfM} & \url{https://github.com/ySalaun/MLSD} \\
		\hline
		\textcolor{blue}{LSD} \cite{LSD,LSDaLineSegmentDetector} & \url{http://www.ipol.im/pub/art/2012/gjmr-lsd/?utm\_source=doi} \\
		\hline
		LSWMS \cite{Linesegmentdetectionusingweightedmeanshiftproceduresona2Dslicesamplingstrategy} & \url{https://sourceforge.net/projects/lswms/} \\
		\hline
		\textcolor{blue}{E2LSD} \cite{E2LSD-J} & \url{https://github.com/roylin1229/E2LSD} \\
		\hline
		\textcolor{blue}{ELSED} \cite{ELSED} & \url{https://iago-suarez.com/ELSED} \\
		\hline
		\textcolor{blue}{FLD} \cite{Outdoorplacerecognitioninurbanenvironmentsusingstraightlines} & \url{https://opencv.org/} \\
		\hline
		CannyLines \cite{CannyLines} & \url{https://cvrs.whu.edu.cn/cannylines/} \\
		\hline
		\textcolor{blue}{EDLines} \cite{EDLines} & \url{https://github.com/CihanTopal/ED\_Lib} \\
		\hline
		AFM \cite{LearningAttractionFieldRepresentationforRobustLineSegmentDetection} & \url{https://github.com/cherubicXN/afm\_cvpr2019} \\
		\hline
		LETR \cite{LineSegmentDetectionUsingTransformerswithoutEdges} & \url{https://github.com/mlpc-ucsd/LETR} \\
		\hline
		\textcolor{blue}{SOLD2} \cite{SOLD2} & \url{https://github.com/cvg/SOLD2} \\
		\hline
		\textcolor{blue}{MLSD} \cite{TowardsRealtimeandLightweightLineSegmentDetection} & \url{https://github.com/navervision/mlsd} \\
		\hline
		\textcolor{blue}{FClip} \cite{FullyConvolutionalLineParsing} & \url{https://github.com/Delay-Xili/F-Clip} \\
		\hline
		TP-LSD \cite{TP-LSD} & \url{https://github.com/Siyuada7/TP-LSD} \\
		\hline
		\cite{Hole-robustWireframeDetection} & \url{https://github.com/SamsungLabs/hole-robust-wf} \\
		\hline
		\cite{LearningtoReconstruct3DManhattanWireframesFromaSingleImage} & \url{https://github.com/zhou13/shapeunity} \\
		\hline
		\textcolor{blue}{LCNN} \cite{End-to-EndWireframeParsing} & \url{https://github.com/zhou13/lcnn} \\
		\hline
		PPGNet \cite{PPGNet} & \url{https://github.com/svip-lab/PPGNet} \\
		\hline
		\textcolor{blue}{HAWP} \cite{Holistically-AttractedWireframeParsing,HAWP} & \url{https://github.com/cherubicxn/hawp} \\
		\hline
		\cite{LearningtoParseWireframesinImagesofMan-MadeEnvironments} & \url{https://github.com/huangkuns/wireframe} \\
		\hline
		\textcolor{blue}{MCMLSD} \cite{MCMLSD} & \url{https://www.elderlab.yorku.ca/mcmlsd/} \\
		\hline
		\textcolor{blue}{HTLCNN} \cite{DeepHoughTransformLinePriors} & \url{https://github.com/yanconglin/Deep-Hough-Transform-Line-Priors} \\
		\hline
		LSDNet \cite{LSDNet} & \url{https://github.com/iitpvisionlab/LSDNet} \\
		\hline
		\textcolor{blue}{DeepLSD} \cite{DeepLSD} & \url{https://github.com/cvg/DeepLSD} \\
		\hline
	\end{tabular}%
	\label{line_detection_implementation}
\end{table}

Local-based approaches generally have a high detection efficiency. Particularly, edge fitting-based approaches \cite{EDLines,ELSED} have the lowest computational costs compared to other approaches since they only consider the drawn edge points instead of whole images when detecting line segments. Local-based methods can exploit local connectivity well to detect line segments and are mathematically explainable. However, most methods in this category only consider local connectivity for line segment detection and lack adequate semantic and global feature constraints. Even with the help of additional aggregation and merging processes, the detected line segments tend to be line fragments instead of long or semantic line segments. The stability of image gradients or edges deeply affects the detection performance, which may be affected by image noise and occlusion. They are suitable for vision applications with low computational and memory costs and low semantic feature requirements, such as VO \cite{8691513} and V-SLAM \cite{9521742}.

Learning-based approaches can achieve line segment detection end-to-end. They can fuse local and global image features for semantic line segment detection. However, the detection performance is deeply affected by the prepared training data. The generalization ability of trained models is difficult to ensure since most are trained based on the Wireframe dataset \cite{LearningtoParseWireframesinImagesofMan-MadeEnvironments}, which only contains 5,000 training images. Furthermore, they are difficult to explain mathematically and to achieve real-time detection without GPU devices, significantly limiting their applications, particularly those with resource-constrained systems and platforms. They are suitable for vision applications with high requirements for semantic understanding, such as image partitioning \cite{7298931} and scene parsing \cite{6619245}.

Hybrid-based approaches generally lack a universal paradigm because they rely on fusing multiple dedicated mechanisms to detect line segments. Consequently, their advantages, disadvantages, and typical applications depend on specific fusion mechanisms. These approaches provide a high degree of flexibility for customization.

\subsection{Challenges and Insights for Potential Solutions}
\label{subsec_detection_challenges}
Global Hough-based approaches have already been well studied, but recently, research on the other three categories is ongoing. Therefore, in this subsection, the challenges for the other three method categories are decisively discussed, and corresponding insights for potential solutions dealing with these challenges are presented.

As mentioned in subsection \ref{subsec_generic_challenges}, the first challenge for line segment detection is that their endpoints are generally unstable due to many factors, such as image noise, occlusion, and illumination variance. An approach for stably determining the line segment endpoints is finding the anchor points on the line segment, such as corners \cite{SODC, ACRA, FASTER} and junctions \cite{End-to-EndWireframeParsing}, which are highly discriminative and can be viewed as stable endpoints for line segments.

For local-based methods, the detected line segments tend to be line segment fragments instead of complete line segments due to the sensitivity to image noise or occlusion. In addition to the existing postprocessing approaches, such as aggregation and merging, as mentioned above, partial global features, \textit{e.g.}, the Hough transformation of limited sample points on line segments, can guide the line segment detection process. In this way, except for local connectivity, partial global features on line segments can be further exploited to detect line segments.

Learning-based methods generally have high computational costs and are deeply affected by the training and testing datasets. Two approaches are presented to potentially address the two challenges. (1) In addition to designing the lightweight network, as in \cite{TowardsRealtimeandLightweightLineSegmentDetection}, combining a shadow network and handcrafted features is feasible for accelerating the line segment detection algorithm inspired by \cite{KeyNet}. (2) In addition to self-supervised \cite{SOLD2} and semisupervised \cite{Hole-robustWireframeDetection} learning schemes, an unsupervised learning scheme can also be considered for line segment detection tasks to decrease the dependency on training and testing datasets with the ground truth inspired by point features \cite{R2D2}.

\section{Line Segment Description Review}
\label{sec_line_description}
This section reviews numerous line segment description approaches. Similar to point \cite{SIFT}, corresponding descriptors can also describe line segments to match them. These descriptors should have various invariant properties, such as rotation, scale, viewpoint, and illumination invariance properties. In addition, line segments contain more structural and geometric features about the scene than point features. Their attributes, \textit{e.g.}, the length and direction, have certain geometric features by themselves and can be utilized to encode line segments. Furthermore, other line segments around the inspected line segment can be exploited to describe line segments structurally by encoding the relative relationship among them.

As introduced in subsection \ref{subsec_high_level_taxonomy}, these description methods can be roughly classified into three categories: (1) statistic-based methods with subcategories of gradient statistic-based and intensity statistic-based approaches (subsection \ref{subsec_description_statistic}), (2) structure-based methods with subcategories of local line segment-assisted and line intersection context-assisted approaches (subsection \ref{subsec_description_structure}), and (3) learning-based methods with line ROS-based and nonline ROS-based subcategories (subsection \ref{subsec_description_learning}). Most line segment descriptors are float descriptors, except for a few methods like \cite{IILB,Abinaryrobustlinedescriptor} and the binary version of \cite{LBD} in OpenCV.

These description methods in three categories have advantages, disadvantages, and positive application scenarios, as listed in TABLE \ref{line_segment_description_comparsion} and analyzed in subsection \ref{subsec_description_comparison}. The insights for potentially addressing the challenges in existing description methods are presented to inspire researchers, as detailed in subsection \ref{subsec_description_challenges}. TABLE \ref{line_description_implementation} lists some open-source implementations of line segment description methods.

\subsection{Statistic-based Approaches}
\label{subsec_description_statistic}

Generally, statistic-based approaches describe line segments by counting the features in line ROSs. Depending on the source of the features, they can be further classified into gradient statistic-based and intensity statistic-based approaches using gradients and intensities, respectively.

\subsubsection{Gradient Statistic-based Approaches}
\label{subsec_description_statistic_gradient}

\begin{figure}[tbp]
	\centering
	\includegraphics[width=0.49\textwidth]{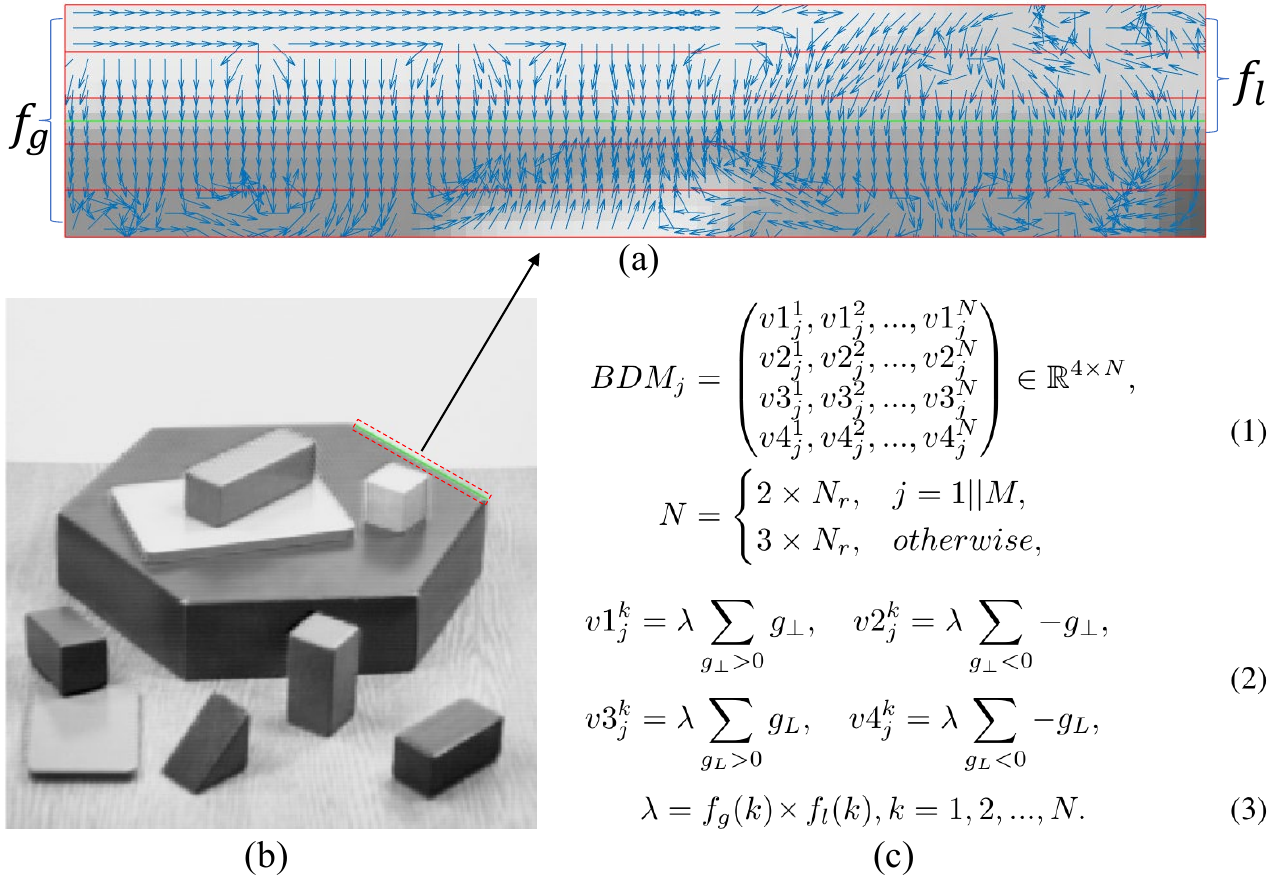}
	\caption{Illustrations of the LBD descriptor \cite{LBD}. (a) The blue arrows represent local gradient vectors with elements parallel ($g_L$) and orthogonal ($g_\bot$) to the line directions, and the red rectangles are local bands in line ROSs. $f_l$ and $f_g$ are weighted Gaussian functions. (b) A line segment (green) with a line ROS (red) in the test image. (c) The band description matrix $BDM_j$ is formulated by counting the gradients from band $B_j$ and its two neighboring bands $B_{j-1}$ and $B_{j+1}$, in which $M$ and $N_r$ are the number of bands and the row number in each band, respectively. The LBD descriptor is formulated by connecting the mean values and standard deviations of various BDMs.}
	\label{lbd_demo}
\end{figure}

\textbf{LBD Descriptor} The local band descriptor (LBD) \cite{LBD} is the baseline method for line segment description. As shown in Fig. \ref{lbd_demo} (a), in the LBD \cite{LBD} approach, the local gradients parallel and orthogonal to the line directions are first calculated and recorded as $g_L$ and $g_\bot$, respectively. Then, $g_L$ and $g_\bot$ are counted in the corresponding bands to formulate the band description matrix (BDM), as shown in Fig. \ref{lbd_demo} (c). Two Gaussian functions, $f_g$ and $f_l$, are used to enhance the LBD descriptor performance. The LBD descriptor is formulated by connecting the mean values and standard deviations of various BDMs and is normalized to improve the robustness in terms of the illumination variance.

\textbf{Related Studies}:
The LBD descriptor was partially inspired by the descriptor \cite{MSLD}, in which the authors exploited image gradients in local subregions to formulate line segment descriptors inspired by the SIFT \cite{SIFT} feature for points. They also presented another descriptor by exploiting the rotation invariant Harris feature \cite{HLD}, generated from image gradients, for line segment description. Liu et al. extended well-known feature point descriptors to line segment descriptions \cite{Extendpointdescriptorsforlinecurveandregionmatching} since line segments can be viewed as a series of points. To achieve the real-time extraction of line segment descriptors, the authors in \cite{FastLineDescriptionforLinebasedSLAM} utilized a constant number of sampled points in line ROSs to calculate gradients and formulate corresponding line descriptors inspired by \cite{MSLD}. The authors in \cite{Matchingofstraightlinesegmentsfromaerialstereoimagesofurbanareas} adapted the Daisy descriptor \cite{DAISY} for feature points to line segments, in which the histogram of gradients is used to construct descriptors. Verhagen et al. proposed a multiple-scale extension for \cite{MSLD}. Benefiting from computing the band difference of image intensity and gradient, an illumination-insensitive binary descriptor \cite{IILB} for line segments was proposed to match line segments under drastic illumination changes.

\subsubsection{Intensity Statistic-based Approaches}
\label{subsec_description_statistic_intensity}
\begin{figure}[tbp]
	\centering
	\includegraphics[width=\linewidth]{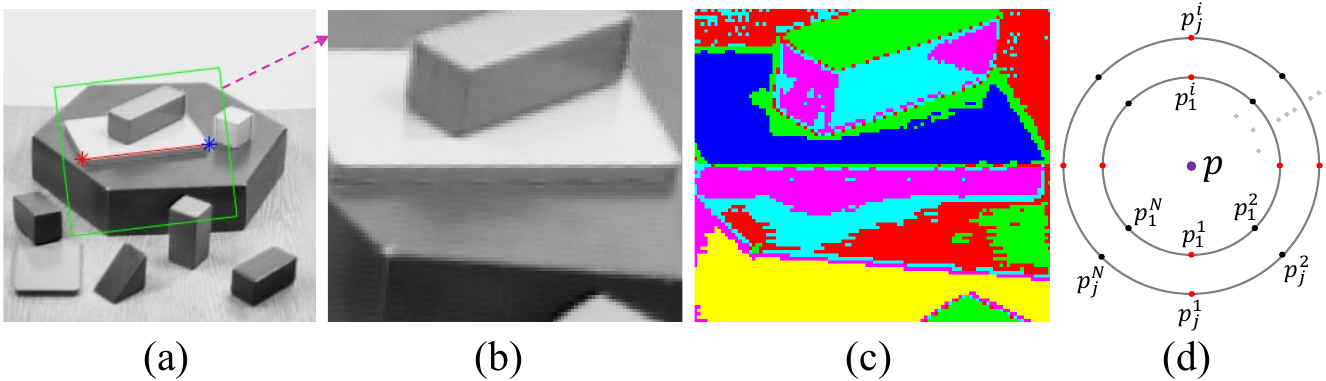}
	\caption{Visual illustrations of the MLIOR descriptor \cite{ALineMatchingMethodBasedonMultipleIntensityOrderingwithUniformlySpacedSampling}. (a) A test line segment with start (red star) and end (blue star) points and its line ROS (green rectangle). (b) Line ROS. (c) Irregular subregions in the line ROS. (d) Sampled points around the target point $p$, where the point set with the same color and circle is grouped to construct the local intensity order representation (LIOR). The MLIOR descriptor is formulated by counting multiple LIORs inspired by \cite{LIOP}.}
	\label{mlior_demo}
\end{figure}

\textbf{MLIOR Descriptor} The multiple local intensity order representation (MLIOR) \cite{ALineMatchingMethodBasedonMultipleIntensityOrderingwithUniformlySpacedSampling} is a representative intensity statistic-based approach. As shown in Fig. \ref{mlior_demo}, different from other methods \cite{LBD,MSLD} using the rectangles to construct subregion divisions, the subregions in line ROSs are formulated based on their intensity order, resulting in an irregular geometry division of subregions. This irregular subregion division makes the MLIOR descriptor insensitive to illumination changes to some extent. For each point in the subregion, the multiple local intensity ordering generated from sampled points around the target point is utilized to describe line segments inspired by \cite{LIOP}. To improve the computational efficiency and decrease the descriptor size, the sampled points are assigned to different circles and point sets to formulate groups when constructing the LIOR.

\textbf{Related Studies}
In addition to the MLIOR descriptor, some other intensity statistic-based line segment descriptors have been proposed. Bay et al. presented a simple line descriptor using a color histogram in HSV color space \cite{Widebaselinestereomatchingwithlinesegments}, in which the intensity values from different image channels are used to describe line segments. The multiple-scale version of \cite{Widebaselinestereomatchingwithlinesegments} was presented in \cite{Scaleinvariantlinedescriptorsforwidebaselinematching}. The authors in \cite{Twoviewlinematchingalgorithmbasedoncontextandappearanceinlowtexturedimages} designed a mixed line segment descriptor, in which the descriptor for reflecting local appearance was formulated based on intensity averages and correlations. Jing et al. proposed a line segment descriptor belonging to both gradient and intensity statistic-based descriptors, in which the gradient magnitude and orientation were counted according to the intensity order of the sampled points \cite{Arobustlinematchingmethodbasedonlocalappearancedescriptorandneighboringgeometricattributes}. 

\subsubsection{Analysis and Summary}
\textbf{\textit{The core idea of statistic-based line segment description approaches lies in exploiting the statistic features, \textit{e.g.}, the histogram of gradients or pixel intensities in line ROSs, to encode line segments.}} Therefore, counting the features, including the subregion division and feature encoding in line ROS, is critical. The subregion division can be regular, as in \cite{LBD}, or irregular, as in \cite{ALineMatchingMethodBasedonMultipleIntensityOrderingwithUniformlySpacedSampling}. A regular division approach is significantly more efficient than an irregular approach because the latter needs to sort the pixels in line ROSs, which is  time-consuming. The subregion division approach determines how the local features will be processed and organized. Feature encoding is performed to explore the invariant characteristics of line segment descriptors. Both of them control the description ability of line segments.

\subsection{Structure-based Approaches}\label{subsec_description_structure}
Structure-based approaches employ additional local line segments to structurally describe their relative relationship. These approaches can be classified into two subcategories based on the usage of local line segments: (1) local line segment-assisted and (2) line intersection context-assisted.

\subsubsection{Local Line Segment-assisted Approaches} \label{subsec_description_structure_local_line} The methods in this subcategory, \textit{i.e.}, \cite{WidebaselineimagematchingusingLineSignatures,Robustaffineinvariantlinematchingforhighresolutionremotesensingimages,Anovelsimilarityinvariantlinedescriptorforgeometricmapregistration,NovelSimilarityInvariantLineDescriptorandMatchingAlgorithmforGlobalMotionEstimation,Twoviewlinematchingalgorithmbasedoncontextandappearanceinlowtexturedimages,AutomaticRegistrationMethodforOpticalRemoteSensingImageswithLargeBackgroundVariationsUsingLineSegments,MultimodalImageRegistrationWithLineSegmentsbySelectiveSearch}, encode local line segments around the inspected line segment to formulate corresponding description signatures. The relative geometric and structural relationships, such as the angle between line segments \cite{Anovelsimilarityinvariantlinedescriptorforgeometricmapregistration,NovelSimilarityInvariantLineDescriptorandMatchingAlgorithmforGlobalMotionEstimation}, the angle between line segment endpoints and line segments \cite{Anovelsimilarityinvariantlinedescriptorforgeometricmapregistration,NovelSimilarityInvariantLineDescriptorandMatchingAlgorithmforGlobalMotionEstimation}, the histogram of local line segments \cite{AutomaticRegistrationMethodforOpticalRemoteSensingImageswithLargeBackgroundVariationsUsingLineSegments}, and the distance between line segment midpoints \cite{Twoviewlinematchingalgorithmbasedoncontextandappearanceinlowtexturedimages}, have been exploited to describe the invariance among these local line segments concerning the inspected line segment.

\subsubsection{Line Intersection Context-assisted Approaches} \label{subsec_description_structure_context} The methods in this subcategory, \textit{i.e.}, \cite{Anovellinematchingmethodbasedonintersectioncontext,Simultaneouslinematchingandepipolargeometryestimationbasedontheintersectioncontextofcoplanarlinepairs,LineMatchinginWideBaselineStereoATopDownApproach,Widebaselinestereomatchingbasedonthelineintersectioncontextforrealtimeworkspacemodeling,RobustLineMatchingBasedonRay-Point-RayStructureDescriptor,HierarchicalLineMatchingBasedonLineJunctionLineStructureDescriptorandLocalHomographyEstimation,Linesegmentmatchingandreconstructionviaexploitingcoplanarcues}, generally find the coplanar and adjacent line segment pairs to formulate the intersecting point, which can be encoded using the well-known feature descriptors for points. The corresponding line segment matches can be recovered based on the matching results of intersecting points. Therefore, the methods in this subcategory transform the line segment description into an intersecting point description.

\subsubsection{Analysis and Summary}
\textbf{\textit{The core idea of structure-based line segment description approaches lies in leveraging additional line segments to describe their relative relationship structurally}}. Therefore, selecting local line segments and encoding the relative relationship among them is critical. Many of them require that the local line segments be coplanar with the inspected line segments since the relative relationship among them is only preserved in coplanar scenarios, which also facilitates the calculation of local homography matrices to recover the precise line-to-line segment correspondences. The methods in this category deeply depend on the line segment detection stability since the local structure is formulated based on line segment detection results.

\subsection{Learning-based Approaches}
\label{subsec_description_learning}
To date, no widely used training and testing datasets exist for learning-based description methods. Most existing methods develop line segment descriptors based on their prepared datasets with the ground truth of line segment matches, either automatically or manually. In addition, their network design schemes are different. However, most of them utilize the triplet loss \cite{NIPS2017_831caa1b} or its variants in the loss function, in which a matched and unmatched pair are exploited. According to the dependence of line ROSs, they can be further classified into two subcategories: line ROS-based approaches and nonline ROS-based approaches.

\subsubsection{Line ROS-based Approaches} \label{subsec_description_learning_ros} Line ROS-based approaches \cite{Abinaryrobustlinedescriptor,DLD,WLD,L2D2,Towardslearninglinedescriptorsfrompatchesanewparadigmandlargescaledataset} train a network by focusing on the local image feature around the inspected line segments. The authors in \cite{DLD} first utilized deep learning to train a network for line segment description by exploiting the triplet loss among three lines, including a matching pair and another nonmatching line, and the approach was improved in their subsequent work \cite{WLD}. Abdellali et al. trained a network to describe line segments in a self-supervised manner with a normalized line ROS size of $32 \times 48$ \cite{L2D2}. The authors in \cite{Towardslearninglinedescriptorsfrompatchesanewparadigmandlargescaledataset} constructed a large-scale dataset of ~229,000 labeled matched line pairs and learned line descriptors from line patches (ROSs).

\subsubsection{Nonline ROS-based Approaches} \label{subsec_description_learning_non_ros} Nonline ROS-based approaches \cite{LearnableLineSegmentDescriptorforVisualSLAM,SOLD2,Superline,LineasaVisualSentenceContextAwareLineDescriptorforVisualLocalization,ELSD,LDAM} can leverage both local image features around the inspected line segment and global image features to encode line segments by using the features in multiple layers. The authors in \cite{LearnableLineSegmentDescriptorforVisualSLAM} designed a lightweight neural network to describe line segments using an automatically collected dataset of matching and nonmatching line segments. Inspired by \cite{SuperPoint}, Pautrat et al. proposed a single deep network for joint line segment detection and description \cite{SOLD2}. The authors in \cite{SOLD2}, \cite{Superline}, and \cite{ELSD} also presented schemes to jointly detect and describe line segments. Yoon et al. presented a context-aware line segment descriptor by dynamically attending to well-describable points on a line \cite{LineasaVisualSentenceContextAwareLineDescriptorforVisualLocalization}.

\subsubsection{Analysis and Summary}
\textbf{\textit{The core idea of learning-based line segment description approaches lies in learning the line segment representations locally or globally based on designed networks with loss functions and prepared training datasets.}} Therefore, designing the network and preparing the training and testing datasets are critical. Like learning-based line segment detection methods, the generalization ability of trained networks needs to be further validated since most of them trained their networks using their prepared datasets. Generally, the descriptors in this category are more robust than these nonlearning-based approaches since they can learn the deep features in multiple layers of the designed network. Most of these methods depend on a GPU for training and predicting. A few of them can extract line segment descriptors in real-time with the help of a powerful GPU.

\subsection{Comparing Methods Across Categories}
\label{subsec_description_comparison}

\begin{table*}[tbp] 	
	\centering 	
	\scriptsize 	
	\caption{Qualitative comparisons for three categories of line segment description algorithms.}		 	\begin{tabular}{|p{0.05\textwidth}<{\centering}|p{0.08\textwidth}<{\centering}|p{0.30\textwidth}<{\centering}|p{0.28\textwidth}<{\centering}|p{0.17\textwidth}<{\centering}|} 		
		\hline 		\textbf{Category}       & \textit{\textbf{e.g.}} & \textbf{Advantages}                                                                                                                                  & \textbf{Disadvantages}                                                                                                                      & \textbf{Typical Applications}                                \\ \hline 		Statistic-based    & LBD \cite{LBD}, MSLD \cite{MSLD}         & (1) Robust to image variance, (2) Mathematically explainable. & (1) Sensitive to low textures, (2) Fewer global and semantic features.                    & Visual localization \cite{PoseEstimationfromLineCorrespondencesACompleteAnalysisandaSeriesofSolutions}, Visual SLAM \cite{9521742} \\ \hline 		Structure-based & NLD \cite{NovelSimilarityInvariantLineDescriptorandMatchingAlgorithmforGlobalMotionEstimation}, LS \cite{WidebaselineimagematchingusingLineSignatures}           & (1) Structure and geometry preserved, (2) Relatively insensitive to low textures, (3) Mathematically explainable.                                                  & (1) Deeply affected by other local points or line segments, (2) Coplanar hypothesis, (3) Fewer global features.           & Map registration \cite{Anovelsimilarityinvariantlinedescriptorforgeometricmapregistration,NovelSimilarityInvariantLineDescriptorandMatchingAlgorithmforGlobalMotionEstimation}, Remote sensing image matching \cite{Robustaffineinvariantlinematchingforhighresolutionremotesensingimages} \\ \hline 		Learning-based & DLD \cite{DLD}, WLD \cite{WLD}              & (1) End-to-end manner, (2)   Fusing local and global features for semantic line segment description.                                      & (1) Deeply affected by training dataset, (2) Difficult to explain mathematically, (3) High computational costs, requiring GPUs. &  Scene parsing \cite{6619245}, 3D scene abstraction \cite{HOFER2017167}                 \\ \hline 	
	\end{tabular} 	
	\label{line_segment_description_comparsion} 
\end{table*}

These description methods in different categories have advantages, disadvantages, and positive applications, as listed in TABLE \ref{line_segment_description_comparsion}. TABLE \ref{line_description_implementation} lists some open-source implementations of line segment description methods. Statistic-based approaches describe line segments by counting the gradients or intensities in line ROSs. Generally, they have a higher computational efficiency than learning-based methods and are insensitive to image variance. They can be explained well mathematically. Most of these approaches are formulated based on line ROSs, leading to insufficient global and semantic features and sensitivity to low-texture scenarios. These approaches are suitable for applications such as visual localization \cite{PoseEstimationfromLineCorrespondencesACompleteAnalysisandaSeriesofSolutions} and SLAM \cite{FastLineDescriptionforLinebasedSLAM}.

\begin{table}[tbp] 
	\centering 
	\scriptsize 
	\caption{Open-source implementations of sampled line segment description methods, in which the methods colored with blue are compared in this review.} 
	\begin{tabular}{|c|c|} 
		\hline 
		\textbf{Method} & \textbf{URL} \\ 
		\hline 
		\textcolor{blue}{IILB} \cite{IILB} & \url{https://github.com/roylin1229/IILB\_descriptor} \\ \hline
		\textcolor{blue}{LBD} \cite{LBD} & \url{https://github.com/mtamburrano/LBD\_Descriptor} \\ \hline
		MSLD \cite{MSLD} & \url{https://kailigo.github.io/projects/LineMatchingBenchmark} \\ \hline
		SMSLD \cite{Scaleinvariantlinedescriptorsforwidebaselinematching} & \url{https://github.com/bverhagen/SMSLD} \\ \hline 
		LJL \cite{HierarchicalLineMatchingBasedonLineJunctionLineStructureDescriptorandLocalHomographyEstimation} & \url{https://cvrs.whu.edu.cn/ljllinematcher/} \\ \hline
		LS \cite{WidebaselineimagematchingusingLineSignatures} & \url{https://kailigo.github.io/projects/LineMatchingBenchmark} \\ \hline 
		\textcolor{blue}{SOLD2} \cite{SOLD2} & \url{https://github.com/cvg/SOLD2} \\ \hline
		\textcolor{blue}{LineTR} \cite{LineasaVisualSentenceContextAwareLineDescriptorforVisualLocalization} & \url{https://github.com/yosungho/LineTR} \\ \hline
		LLD \cite{LearnableLineSegmentDescriptorforVisualSLAM} & \url{https://github.com/alexandervakhitov/lld-public} \\ \hline
		\textcolor{blue}{DLD} \cite{DLD} & \url{https://github.com/manuellange/DLD} \\ \hline
		\textcolor{blue}{WLD} \cite{WLD} & \url{https://github.com/manuellange/WLD} \\ \hline 
	\end{tabular}%
	\label{line_description_implementation} 
\end{table}

Structure-based approaches describe line segments by leveraging additional line segments in the neighborhood. They can encode the structural and geometric features around the inspected line segment. Therefore, they are essentially structurally and geometrically preserved and insensitive to low textures to an extent. Similar to statistic-based approaches, they can also be well explained mathematically. However, since they use additional line segments to encode line segments, they are deeply affected by the line segment detection performance. In addition, most of these approaches require the precondition of the coplanar hypothesis, limiting their application to some extent. They can exploit the larger neighborhood context than statistic-based approaches to encode line segments but also lack corresponding global features. They are suitable for applications such as map registration \cite{NovelSimilarityInvariantLineDescriptorandMatchingAlgorithmforGlobalMotionEstimation}.

Learning-based approaches describe line segments using neural networks end-to-end, in which both the local and global features of line segments can be learned. However, training and testing data deeply affect their performance, and it is not easy to know whether they will perform well on other evaluation data since most of these networks are trained using their prepared datasets. In addition, they generally have difficulty achieving real-time line segment descriptor extraction without the help of powerful GPU devices and are difficult to explain mathematically. They are suitable for offline and complex applications such as scene parsing \cite{6619245} and 3D scene abstraction \cite{HOFER2017167}.

\subsection{Challenges and Insights for Potential Solutions}
\label{subsec_description_challenges}
As mentioned in subsection \ref{subsec_generic_challenges}, the first challenge for line segment description mentioned above also results from the unstable endpoints of detected line segments since the corresponding line ROSs or the local structure around the inspected line segment are generated according to the line segment endpoints. A new insight for dealing with this challenge lies in describing line segments in the Hough space, where the line segment endpoints are insensitive.

The second challenge lies in the lack of binary descriptors for line segments, limiting their usage in applications with high requirements of computational and storage costs. Except for the approaches in \cite{Abinaryrobustlinedescriptor,IILB} and the binary version of \cite{LBD}, almost all the line segment descriptors are float descriptors. Binary descriptors have significant advantages over float descriptors regarding computational and storage costs. A straightforward method for addressing this challenge is to design binary line segment descriptors based on well-developed binary point feature descriptors.

The third challenge is the inability to describe line segments in challenging scenarios, as mentioned in subsection \ref{subsec_generic_challenges}. To address this issue, some methods can be considered.
\begin{itemize}
	\item Homography and epipolar geometry constraints can be employed to enhance the line segment matching process, as in \cite{Automaticlinematchingacrossviews,MatchingDisparateViewsofPlanarSurfacesusingProjectiveInvariants,Thegeometryandmatchingoflinesandcurvesovermultipleviews,Anovellinematchingmethodbasedonintersectioncontext,Simultaneouslinematchingandepipolargeometryestimationbasedontheintersectioncontextofcoplanarlinepairs,LineMatchinginWideBaselineStereoATopDownApproach,Widebaselinestereomatchingbasedonthelineintersectioncontextforrealtimeworkspacemodeling,RobustLineMatchingBasedonRay-Point-RayStructureDescriptor,NovelCoplanarLinePointsInvariantsforRobustLineMatchingAcrossViews,HierarchicalLineMatchingBasedonLineJunctionLineStructureDescriptorandLocalHomographyEstimation,Linesegmentmatchingandreconstructionviaexploitingcoplanarcues,MultimodalImageRegistrationWithLineSegmentsbySelectiveSearch,Robustlinesegmentmatchingviareweightedrandomwalksonthehomographygraph}. These geometry constraints can be used when corresponding requirements, such as the coplanar hypothesis or image pose existence, are satisfied.
		
	\item For vision tasks operating in videos, line segment tracking \cite{Trackinglinesegments,GeometricbasedLineSegmentTrackingforHDRStereoSequences,LOF,LineFlowBasedSimultaneousLocalizationandMapping,HighlyEfficientLineSegmentTrackingwithanIMUKLTPredictionandaConvexGeometricDistanceMinimization,SparseOpticalFlow-BasedLineFeatureTracking} can also be exploited to enhance the line segment matching process.
		
	\item Additional point features can be used to enhance the capability of line segment description \cite{MatchingDisparateViewsofPlanarSurfacesusingProjectiveInvariants,Linematchingleveragedbypointcorrespondences,Robustlinematchingthroughlinepointinvariants,Scaleinvariantlinematchingonthesphere,Line-sweep,NovelCoplanarLinePointsInvariantsforRobustLineMatchingAcrossViews,Dude,Linematchingbasedonlinepointsinvariantandlocalhomography,Linematchingbasedonplanarhomographyforstereoaerialimages}. These methods exploit the local coplanar points and lines to describe line segments by leveraging the intrinsic invariance among them, such as the projective invariant method that uses two lines and two points \cite{MatchingDisparateViewsofPlanarSurfacesusingProjectiveInvariants} and the affine invariant approach that uses one line and two points \cite{Linematchingleveragedbypointcorrespondences, Robustlinematchingthroughlinepointinvariants}. In recent studies \cite{HDPL,GlueStick}, deep neural networks have been specifically crafted to describe and match point features and line segments concurrently.
	
	\item In addition to the commonly employed matching strategy outlined in \cite{SIFTMatchingbyContextExposed}, specialized matching methodologies, such as those proposed in \cite{WGLSM, SOLD2,GlueStick,HDPL}, can be developed to mitigate limitations in line segment description to some extent.
\end{itemize}

\section{Performance Analysis}
\label{sec_performance}

\begin{table*}[tbp]
	\centering
	\setlength\tabcolsep{2pt}
	\scriptsize
	\caption{Testing datasets for evaluating the performance of line segment detection and description methods in this review.}
		\begin{tabular}{|c|c|c|c|p{9cm}|}
			\hline
			\textbf{Dataset} & \textbf{\# Groups/\# Images} & \textbf{Evaluation type} & \textbf{Ground truth} & \textbf{Note} \\ \hline
			HPatches \cite{HPatches} & 116/696 & Detection and description & N/A & Natural images with variations in illumination and viewpoint. \\ \hline
			KADID-10k \cite{KADID} & 81/10,206 & Detection and description & N/A & Testing images with artificial distortions of blur, color distortion, compression, noise, brightness change, spatial distortion, and sharpness/contrast. \\ \hline
			RDNIM \cite{RDNIM} & 17/1,739 & Detection and description & N/A & Natural images with variations in light and homographic warp. \\ \hline
			DNIM \cite{DNIM} & 17/1,722 & Detection and description & N/A & Natural images with variation in light. \\ \hline
			Apollo \cite{IILB} & 1,000/2,087 & Detection and description & N/A & Synthetic images with variation in light. \\ \hline
			VGGaffine \cite{vggaffine} & 8/48 & Detection and description & N/A & Natural images with variations in blur, viewpoint, zoom/rotation, light, and JPEG. \\ \hline
			Wireframe \cite{LearningtoParseWireframesinImagesofMan-MadeEnvironments} & 462/462 & Detection & Wireframe & Natural images in both indoor and outdoor scenarios. \\ \hline
			YorkUrban \cite{ANovelLineletBasedRepresentationforLineSegmentDetection} & 102/102 & Detection & Line segment & Natural images in both indoor and outdoor scenarios. \\ \hline
		\end{tabular}
	\label{eva_datasets}
\end{table*}

In this section, numerous SOTA line segment detection and description methods are qualitatively and quantitatively evaluated to review their performance unbiasedly and strictly. They are evaluated separately because many existing studies individually presented their line segment detection or description methods. Instead of evaluating line segments in specific vision tasks as in \cite{EVOLIN}, the line segment properties mentioned in subsection \ref{subsec_ls_properties} are evaluated in this review. They do not depend on specific vision tasks. Specifically, fifteen line segment detection methods in different categories are compared, as listed in TABLE \ref{line_detection_implementation}.

In addition, six open-source line segment description methods listed in TABLE \ref{line_description_implementation}, along with the binary version of the LBD \cite{LBD} descriptor integrated into OpenCV, the MLIOR descriptor \cite{ALineMatchingMethodBasedonMultipleIntensityOrderingwithUniformlySpacedSampling}, and the GlueStick \cite{GlueStick} matching method, are compared. Except for the IILB descriptor and the binary version of the LBD descriptor, the other testing descriptors are float descriptors. The statistic-based and learning-based methods are also called appearance-based methods in some studies. The structure-based approaches are excluded from the comparison because they require additional line segments; therefore, comparing them with the other two types of methods is unfair.

Note that the related parameters of the compared detection and description methods are set to their default values as suggested in corresponding references or recommended by the open-source implementations. TABLE S1 to S3 in the supplementary material display the implementation/model details for the compared methods.

\subsection{Evaluation Datasets}
\label{subsec_performance_dataset}
As mentioned in subsection \ref{subsec_ls_properties}, line segments should be accurately, robustly, stably, and repeatably detected, described, and matched in multiple images containing the same scene. Regrettably, as outlined in subsection \ref{subsec_generic_challenges}, the absence of gold-standard datasets for assessing line segment detection and description methods is notable. The available datasets typically cover only limited aspects of line segment detection and description performance. Consequently, this review employs multiple datasets to comprehensively evaluate the compared methods. Specifically, instead of using task-specific datasets \cite{EVOLIN,ETH3D,BADSLAM}, six well-known datasets summarized in TABLE \ref{eva_datasets}, \textit{i.e.}, the HPatches \cite{HPatches}, KADID-10k \cite{KADID}, DNIM \cite{DNIM}, RDNIM \cite{RDNIM}, VGGaffine \cite{vggaffine}, and Apollo \cite{IILB} datasets, are employed to evaluate various line segment detection and description algorithms. The corresponding homography matrices enable pixel-to-pixel correspondences between two test images, which in turn provide line segment to line segment correspondences for evaluation. They include test image groups with designed image variance properties.

Furthermore, line segment detection can be regarded as a typical object detection task, in which the carefully and manually labeled ground truth of line segments is employed as a reference for evaluation. Thus, as mentioned in subsection \ref{subsubsec_detection_dataset} and summarized in TABLE \ref{eva_datasets}, two datasets with manually labeled ground truth line segments, \textit{i.e.}, the improved YorkUrban \cite{ANovelLineletBasedRepresentationforLineSegmentDetection} and Wireframe \cite{LearningtoParseWireframesinImagesofMan-MadeEnvironments} datasets, are employed to further evaluate line segment detection methods.

\subsection{Evaluation Metrics}
\label{subsec_performance_metric}
\subsubsection{Metrics for Line Segment Detection Evaluation}


\begin{figure}[tbp]
	\centering
	\includegraphics[width=\linewidth]{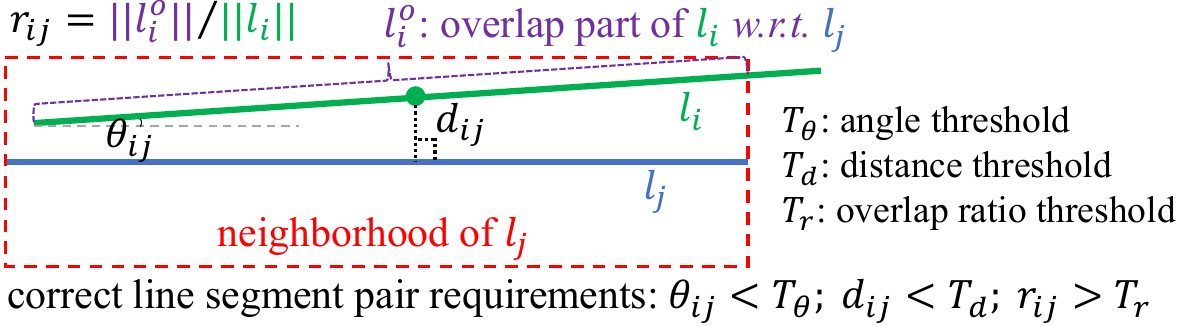}
	\caption{Two line segments $l_i$ and $l_j$ form a correct line segment pair if they mutually overlap the most and if each lies within the neighborhood of the other, as determined by threshold requirements on the distance $d_{ij}$ between the middle point of $l_i$ and $l_j$, the angle difference $\theta_{ij}$ between $l_i$ and $l_j$, and the overlap ratio $r_{ij}$, as illustrated in the figure.}
	\label{matched_line_segment}
\end{figure}

As mentioned above, eight datasets with or without ground truth are utilized to assess various line segment detection algorithms from two dimensions. For the category of nonground truth-based approaches, repeatability $\mathcal{R}$, location error $\mathcal{L}_d$, and orientation error $\mathcal{L}_\theta$, widely used in low-level feature evaluation \cite{CPDA}, are exploited to evaluate the numerous methods. It can effectively indicate the ability to repeatedly and accurately detect line segments in different images containing the same scene. Inspired by \cite{CPDA}, the repeatability in this review is defined as $\mathcal{R} = {N_c}/{2} \times (1/{N^1_t} + 1/{N^2_t})$, in which $N_c$, $N^1_t$, and $N^2_t$ are the numbers of correct line segment pairs and valid line segments in the first and second test images, respectively. The detected line segments in the first image are valid if the projected line segments, according to the homography matrix, lie in the second image and vice versa. In addition, the length of line segments before and after projection should be larger than a threshold, \textit{e.g.}, 15 pixels in this review, to remove line fragments. The definition of correct line segment pairs is explained in Fig. \ref{matched_line_segment}. Location error $\mathcal{L}_d$ and orientation error $\mathcal{L}_\theta$ are also explained by the $d_{ij}$ and $\theta_{ij}$ in Fig. \ref{matched_line_segment}. 

For the ground truth-based approach, the $F_1$ score, formulated based on detection precision $N_c/N_t$ and recall $N_c/N_g$, is adopted as the key metric to evaluate the consistency with the labeled ground truth. It can be defined as $F_1 = 2 \times N_c / (N_g+N_t)$, in which $N_c$, $N_g$, and $N_t$ are the number of correct line segment pairs, line segment ground truth, and detected line segments, respectively. 

In addition, this review compares the runtime performance based on frames per second (FPS) of selected line segment detection methods, providing insights into their detection efficiency.

\subsubsection{Metrics for Line Segment Description Evaluation}
The matching precision ($\mathcal{P}_{m}$) and recall ($\mathcal{C}_{m}$) are defined to evaluate the matching ability of line segment descriptors. They can be formulated as $\mathcal{P}_{m} = {N^c_m}/{N^p_m}$ and $\mathcal{C}_{m} = {N^c_m}/{N^h_m}$. Here, $N^p_m$ represents the number of putative line segment matches between two test images based on descriptor similarity. $N^c_m$ denotes the number of correct matches within $N^p_m$, validated by the corresponding homography matrix. $N^h_m$ is the number of matched line segments between two test images determined by the homography matrix instead of descriptor similarity, as shown in Fig. \ref{matched_line_segment}.

\subsection{Results for Line Segment Detection Evaluation}
\begin{figure*}[tbp]
	\centering
	\includegraphics[width=\textwidth]{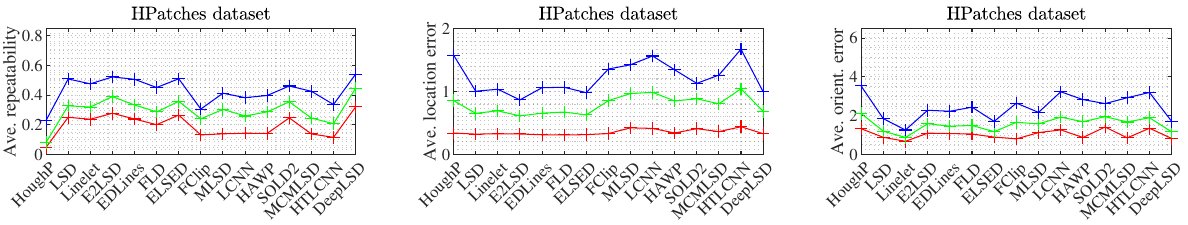}
	\includegraphics[width=\textwidth]{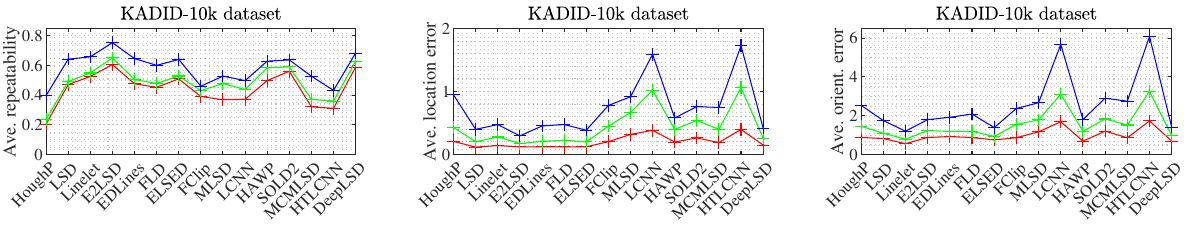}
	\includegraphics[width=\textwidth]{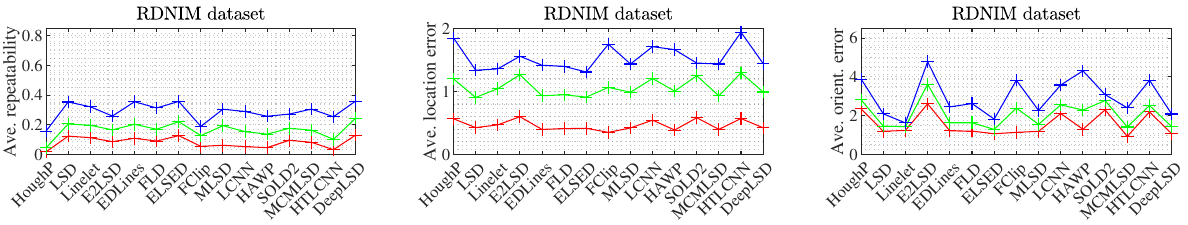}
	\includegraphics[width=0.75\textwidth]{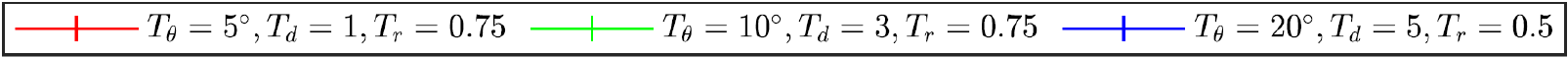}
	\caption{The figure plots the average repeatability scores, location errors, and orientation errors of fifteen testing methods for line segment detection evaluated on the HPatches \cite{HPatches}, KADID-10k \cite{KADID}, and RDNIM \cite{RDNIM} datasets.}
	\label{line_segment_detection_plot}
\end{figure*}

As shown in Fig. \ref{matched_line_segment}, the thresholds of angle $T_\theta$, distance $T_d$, and overlap ratio $T_r$ are three critical evaluation parameters in the experiment. They control the strictness of matched line segments. Here, three configurations are employed in the experiments: (1) $T_\theta = 5^\circ, T_d = 1, T_r = 0.75$ for a strict configuration, (2) $T_\theta = 10^\circ, T_d = 3, T_r = 0.75$ for a moderate configuration, and (3) $T_\theta = 20^\circ, T_d = 5, T_r = 0.5$ for a loose configuration. Additional experimental results with details can be found in the supplementary material.

\subsubsection{Results of the Nonground Truth-based Approach}
Fig. \ref{line_segment_detection_plot} and Fig. S1 to S4 in the supplementary material present the average repeatability scores, location errors, and orientation errors derived from fifteen testing methods assessed across eight evaluation datasets under three evaluation configurations. The results show some interesting observations. (1) No single algorithm consistently outperforms others across all test datasets and evaluation metrics. (2) Despite their typically higher computational demands, learning-based methods do not exhibit significant performance enhancements as anticipated. (3) Global Hough-based approaches yield subpar results. (4) Except for the datasets primarily focusing on variations in light conditions, both local-based methods and the hybrid-based DeepLSD approach generally outperform others.

\subsubsection{Results of the Ground Truth-based Approach}
\begin{figure*}[tbp]
	\centering
	\includegraphics[width=\textwidth]{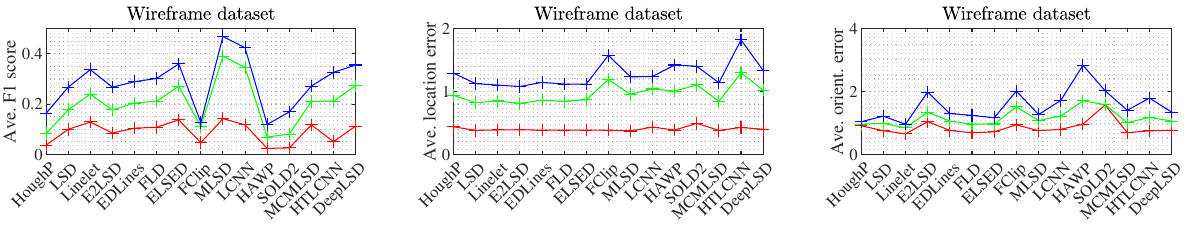}
	\includegraphics[width=\textwidth]{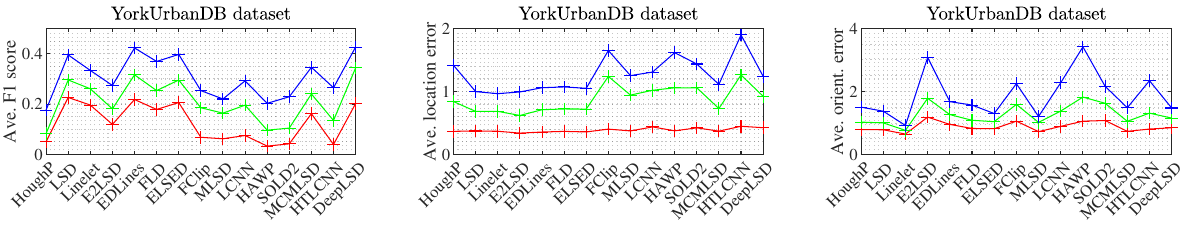}	
	\includegraphics[width=0.75\linewidth]{./figures/detection/bar.pdf}
	\caption{The figure plots the average $F_1$ scores, location errors, and orientation errors of fifteen testing methods for line segment detection evaluated on the Wireframe \cite{LearningtoParseWireframesinImagesofMan-MadeEnvironments} and improved YorkUrban \cite{ANovelLineletBasedRepresentationforLineSegmentDetection} datasets.}
	\label{Wireframe_YorkUrban_detection}
\end{figure*}

Fig. \ref{detection_visualization} and Fig. S5 in the supplementary material provide a qualitative visualization of the line segments detected by six methods across four categories. Fig. \ref{Wireframe_YorkUrban_detection} presents a quantitative analysis of average $F_1$ scores, location errors, and orientation errors for fifteen testing methods evaluated on the Wireframe and YorkUrban datasets under three evaluation configurations. The results reveal several noteworthy observations. (1) Learning-based methods generally exhibit inferior performance in terms of location and orientation errors, particularly under moderate and loose evaluation configurations. (2) In the context of the YorkUrban dataset, local-based approaches and the DeepLSD algorithm \cite{DeepLSD} generally outperform other methods. (3) Global Hough-based approaches demonstrate poor performance.

\begin{figure*}[tbp]
	\centering
	\includegraphics[width=\textwidth]{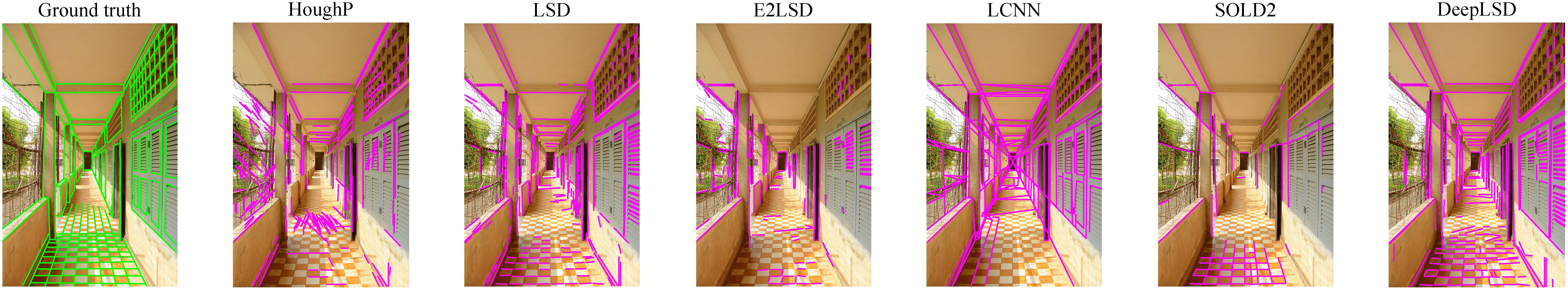}
	\includegraphics[width=\textwidth]{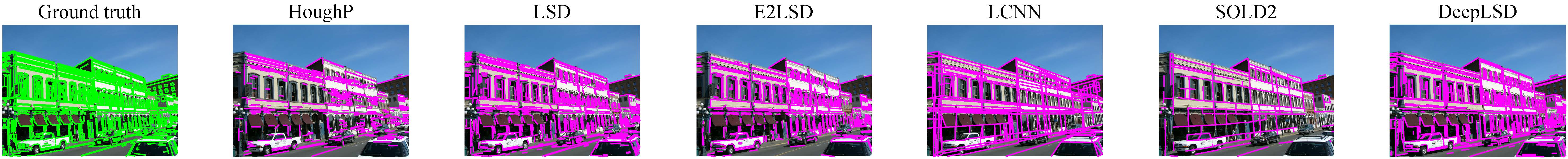}
	\caption{Line segments detected in two test images using six different methods across four categories. The top image is sourced from the Wireframe dataset \cite{LearningtoParseWireframesinImagesofMan-MadeEnvironments}, while the bottom image is from the YorkUrban dataset \cite{ANovelLineletBasedRepresentationforLineSegmentDetection}.}
	\label{detection_visualization}
\end{figure*}

\subsubsection{Results of the Runtime Performance}
TABLE \ref{table_runtime} presents the FPS for eight methods assessed on the Wireframe dataset \cite{LearningtoParseWireframesinImagesofMan-MadeEnvironments}. The results align with the analysis in TABLE \ref{line_segment_detection_comparsion} and subsection \ref{subsec_detection_pro_cons}. Specifically, edge fitting-based methods demonstrate a distinct advantage over other methods, and learning-based approaches often incur significant computational costs, necessitating the utilization of powerful GPU devices for real-time detection implementation.

\begin{table}[tbp]
	\centering
	\setlength\tabcolsep{4pt}
	\scriptsize
	\caption{The FPS of eight line segment detection methods evaluated on the wireframe dataset \cite{LearningtoParseWireframesinImagesofMan-MadeEnvironments}. (CPU: 13th Intel® Core™ i7-13790F; GPU: NVIDIA GeForce RTX 4060 Ti.)}
	\begin{tabular}{|c|c|c|c|c|c|}
		\hline
		\textbf{Method}  & \textbf{Configuration}    & \textbf{FPS} & \textbf{Method}   & \textbf{Configuration}        & \textbf{FPS} \\ \hline
		EDLines \cite{EDLines}    & CPU, OpenCV & 357 & LCNN \cite{End-to-EndWireframeParsing}       & GPU, PyTorch & 2   \\ \hline
		LSD \cite{LSD}      & CPU, OpenCV & 141 & MLSD \cite{TowardsRealtimeandLightweightLineSegmentDetection}       & GPU, PyTorch & 70  \\ \hline
		FLD \cite{Outdoorplacerecognitioninurbanenvironmentsusingstraightlines}       & CPU, OpenCV & 357 & HAWP \cite{HAWP}       & GPU, PyTorch & 37  \\ \hline
		HoughP \cite{RobustDetectionofLinesUsingtheProgressiveProbabilisticHoughTransform}    & CPU, OpenCV & 222 & DeepLSD \cite{DeepLSD}    & GPU, PyTorch & 15  \\ \hline
	\end{tabular}
	\label{table_runtime}
\end{table}

\subsubsection{Interpretations of Evaluation Results}
The following explanations elaborate on the noteworthy findings described above.
\begin{itemize}
	\item In both evaluations, global Hough-based methods exhibit subpar performance. This deficiency primarily stems from their insufficient local connectivity and semantic feature constraints.
	\item Except for the Wireframe dataset and datasets focusing primarily on variations in light conditions, local-based methods and the hybrid-based DeepLSD approach outperform other methods in general. This superiority is attributed to their comprehensive utilization of local connectivity, enabling the detection of line segments with highly consistent local features.
	\item While learning-based approaches may achieve comparable performance on the Wireframe dataset, their effectiveness diminishes on other datasets due to their reliance on training exclusively on this dataset, highlighting their limited generalization capacity.
	\item Learning-based approaches demonstrate inadequate performance regarding location and orientation errors, largely stemming from their insensitivity to accuracy, as discussed in subsection \ref{subsec_detection_learning_analysis}.
\end{itemize}

\subsection{Results for Line Segment Description Evaluation}
\subsubsection{Evaluation with Predefined Line Segments}
The evaluation of line segment descriptions heavily relies on line segment detection, as these descriptors are primarily constructed based on detected line segments. To ensure a rigorous evaluation setting, two strategies are adopted to minimize the influence of line segment detection. First, all descriptors are tested using the same set of line segments detected by the LSD method \cite{LSD}. Second, inspired by \cite{HPatches} to address the inconsistent problem of line segment endpoints, line segments are extracted only from the first test image and then projected onto other test images according to homography matrices for each test group. The mutual brute-force strategy is adopted for all descriptors.

\subsubsection{Evaluation with Line Segment Detection Methods}
Additionally, this review also acknowledges that in practical applications, line segment descriptors are often used in conjunction with line segment detection. Therefore, this review further evaluates line segment description algorithms based on selected line segment detection methods. TABLE S3 in the supplementary material presents specific evaluation details, including description algorithms, detection algorithms, matching algorithms, and implementation specifics.

\subsubsection{Evaluation Results}
Fig. \ref{descriptor_without_detection_performance} and Fig. S6 to S9 in the supplementary material provide quantitative plots of the average matching precision ($\mathcal{P}_m$) and recall ($\mathcal{C}_m$) values for six testing methods utilizing predefined line segments. Fig. S10 in the supplementary material provides a qualitative visualization of the matched line segments using four methods. Fig. \ref{descriptor_performance} and Fig. S11 to S14 in the supplementary material provide quantitative plots of the average matching precision ($\mathcal{P}_m$) and recall ($\mathcal{C}_m$) values for nine testing methods utilizing selected detection approaches. These results reveal several noteworthy observations: (1) Learning-based approaches, despite their higher computational and memory costs, do not consistently outperform statistical-based methods as anticipated. (2) Binary descriptors can exhibit competitive performance despite their smaller descriptor size requirements. (3) The GlueStick \cite{GlueStick} approach generally outperforms other methods.

\begin{figure*}[tbp]
	\centering
	\includegraphics[width=0.245\textwidth]{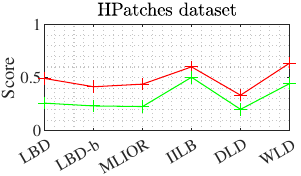}
	\includegraphics[width=0.245\textwidth]{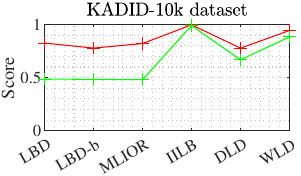}
	\includegraphics[width=0.245\textwidth]{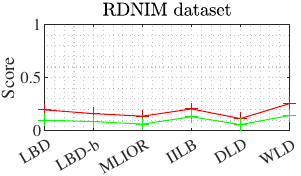}
	\includegraphics[width=0.245\textwidth]{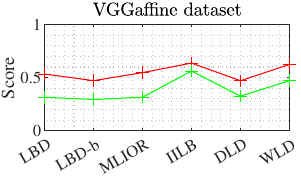}
	\includegraphics[width=0.35\linewidth]{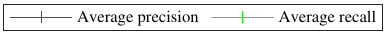} 
	\caption{The figure plots the average matching precision ($\mathcal{P}_m$) and recall ($\mathcal{C}_m$) values of six testing methods with predefined line segments evaluated on the HPatches \cite{HPatches}, KADID-10k \cite{KADID}, RDNIM \cite{RDNIM}, and VGGaffine \cite{vggaffine} datasets.}
	\label{descriptor_without_detection_performance}
\end{figure*}

\begin{figure*}[tbp]
	\centering
	\includegraphics[width=0.245\textwidth]{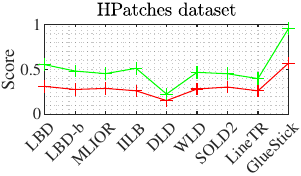}
	\includegraphics[width=0.245\textwidth]{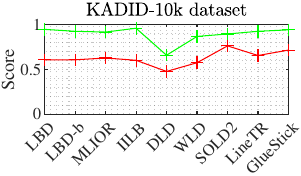}
	\includegraphics[width=0.245\textwidth]{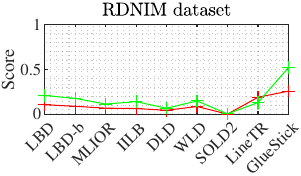}
	\includegraphics[width=0.245\textwidth]{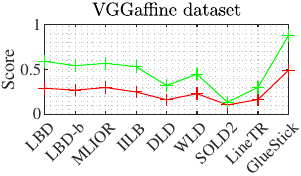}
	\includegraphics[width=0.35\linewidth]{./figures/description/bar.pdf} 
	\caption{The figure plots the average matching precision ($\mathcal{P}_m$) and recall ($\mathcal{C}_m$) values of nine testing methods with selected detection approaches evaluated on the HPatches \cite{HPatches}, KADID-10k \cite{KADID}, RDNIM \cite{RDNIM}, and VGGaffine \cite{vggaffine} datasets.}
	\label{descriptor_performance}
\end{figure*}

\subsubsection{Interpretations of Evaluation Results}
	The following explanations elaborate on the noteworthy findings described above.
\begin{itemize}
\item Regarding observation (1), the network design based on prepared data deeply affects the ability of learning-based approaches. While most of them perform remarkably well in their respective testing datasets, their performance diminishes when applied to other testing datasets. This suggests a constraint on the generalization ability of these models.
\item Regarding observation (2), despite the negative impact of descriptor size on performance, the performance of binary descriptors is predominantly determined by the binary quantization strategy employed.
\item The GlueStick outperforms other methods primarily because it integrates feature point information in both description and matching phases.
\end{itemize}

\section{Future Research Directions}
\label{research_directions}
The previous sections analyzed and summarized existing line segment detection and description algorithms. This section looks at potentially interesting research directions, providing insights for researchers about line segment detection and description.

\begin{itemize}
	\item \textbf{Line Segment Detection and Description in Color Space.}
	Theoretically, low-level image features extracted in color space have more discriminative ability than those in gray space, as mentioned in \cite{1542040}. However, most nondeep learning-based line segment detection and description methods operate on gray images, ignoring the corresponding color features to improve efficiency. With increasing computational resources, the concern can be decreased in the future, particularly those without real-time performance requirements. Thus, nondeep learning-based line segment detection and description in color space may receive more attention in the future.
	
	\item \textbf{Line Segment Detection and Description in the Spatiotemporal Domain}
	Local image features extracted in the spatiotemporal domain can reveal more features than those extracted only in the spatial domain, facilitating many vision tasks such as human action recognition \cite{6725627}. However, many existing line segment detection and description methods generally operate on single images (spatial domain) and ignore relationships in the time domain, limiting their applications for sequential vision tasks. In the future, line segment detection and description methods in the spatiotemporal domain may also receive more attention.
		
	\item \textbf{Cross-Modality Line Segment Detection and Description}
	Recently, multiple sensor configurations have been widely employed in many applications, such as sensor fusing-based localization and mapping \cite{9502143}. The cross-modality low-level features \cite{9678058} can significantly improve the line segment detection performance and description since they can leverage additional data to overcome the drawback of single-sensor data. Therefore, cross-modality line segment detection and description may receive more attention in the future.
		
	\item \textbf{Line Segment Detection and Description in Other Image Formats}
	With the rapid development of image sensing technology, more images, such as panoramic, spherical, infrared, depth, and event images, are available, providing additional and special sensing information compared with regular images. Many point detection and description methods have been applied to these images. However, only a few line segment detection \cite{ULSD,DetectingLineSegmentsinMotionblurredImageswithEvents} and description methods have been applied to these images. Therefore, line segment detection and description for these images may also receive more attention in the future.
		
	\item \textbf{Hardware Architecture Design and Implementation for Line Segment Detection and Description}
	Optimizing the efficiency of line segment detection and description algorithms is important because it forms the basis for numerous vision tasks. Similar to point \cite{8304764,8382255} and edge \cite{9546047} features, line segment detection and description algorithms can also be accelerated using hardware architecture.
\end{itemize}

\section{Conclusion}
\label{sec_conclusion}
This study comprehensively reviewed more than one hundred line segment detection and description algorithms to give academics an overall picture and deep understanding of them. Based on their mechanisms, two taxonomies were presented to introduce, analyze, and summarize these algorithms, facilitating researchers in quickly and extensively learning about them. For the algorithms in each category, their key issues, core ideas, advantages and disadvantages, and potential applications were also analyzed and summarized, including some previously unknown findings. This review also summarized existing challenges in line segment detection and description algorithms and provided some insights to potentially address these challenges. Some representative SOTA methods were unbiasedly evaluated based on natural and synthetic datasets to further show the performance of various line segment detection and description algorithms. The theoretical analysis and experimental results can guide scholars in selecting the best method for their intended applications. Finally, this study proposed potentially interesting future research directions to attract more attention from researchers to this field.

\bibliography{IEEEabrv,line_features,line_features_detection,line_features_description,others}
\bibliographystyle{IEEEtran}

\end{document}